\documentclass[sigconf]{acmart}

\setcopyright{none}
\acmDOI{}
\acmISBN{}
\acmConference[]{}{}{}
\acmBooktitle{}
\acmPrice{}

\definecolor{colA}{RGB}{232,242,252}  
\definecolor{colB}{RGB}{236,250,242}  
\definecolor{colC}{RGB}{252,240,240}  

\colorlet{secondbest}{cyan!15}
\definecolor{colA}{RGB}{232,242,252}
\definecolor{colB}{RGB}{236,250,242}
\definecolor{colC}{RGB}{252,240,240}

\colorlet{secondbest}{blue!15}
\usepackage{graphicx}
\usepackage{subcaption}
\usepackage{adjustbox}
\usepackage{wrapfig}

\usepackage{amsmath,amssymb,amsfonts}
\usepackage{amsthm}
\usepackage{bm}
\usepackage{mathtools}
\usepackage{nicefrac}

\usepackage[inline]{enumitem}
\usepackage{booktabs}
\usepackage{multirow}
\usepackage{makecell}

\usepackage{colortbl}

\usepackage{algorithm}
\usepackage[noend]{algpseudocode}

\usepackage{tikz}
\usetikzlibrary{fit,backgrounds,arrows,decorations.markings,arrows.meta,matrix}
\usepackage{pgfplots}
\usepgfplotslibrary{groupplots}

\usepackage{comment}
\usepackage{relsize}
\usepackage{bbm}
\usepackage{ctable}
\usepackage[symbol]{footmisc}

\newtheorem{theorem}{Theorem}
\newtheorem*{theorem*}{Theorem}
\newtheorem{lemma}{Lemma}
\newtheorem*{lemma*}{Lemma}
\newtheorem{corollary}{Corollary}
\theoremstyle{definition}

\theoremstyle{remark} \newtheorem{remark}{Remark}
\definecolor{colA}{HTML}{B3E5FC} 
\definecolor{colB}{HTML}{C8E6C9} 
\definecolor{colC}{HTML}{FFE0B2} 

\newcommand{\newpair}[1]{\textcolor{blue!70!black}{#1}}
\newcommand{\oldpair}[1]{\textcolor{black!45}{#1}}

\usepackage{longtable}
\usepackage{array}
\usepackage{booktabs}

\title{\textsc{ATLAS}: Adaptive Topology\mbox{-}based Learning at Scale for Homophilic and Heterophilic Graphs}

\author{Turja Kundu}
\affiliation{%
  \institution{University of North Texas}
  \city{Denton}
  \state{Texas}
  \country{USA}
}
\email{turjakundu@my.unt.edu}

\author{Sanjukta Bhowmick}
\affiliation{%
  \institution{University of North Texas}
  \city{Denton}
  \state{Texas}
  \country{USA}
}
\email{sanjukta.bhowmick@unt.edu}

\begin{document}

\newpage

\begin{abstract}
Graph neural networks exploit structure through iterative message
passing. The process is expensive and works best when a node
shares its neighbors' class, i.e.\ the graph is homophilous. For
heterophilous graphs, most methods focus on modifying the topology
as a signal to be smoothed when, in fact, the classification is
driven by several complementary channels operating at different
scales. We present ATLAS (Adaptive Topology-based Learning at
Scale), a framework with propagation-free training and
adjacency-free inference after one-time preprocessing, which
extracts topology as explicit, precomputed channels, specifically
multi-resolution community structure, one-hop neighbor attributes,
and multi-hop label priors, and trains a compact MLP on the
augmented node representation. Every channel is computed once during
preprocessing, so training needs no neighborhood expansion and
inference touches no adjacency matrix. We show theoretically that
given finite-sample data, each channel obeys an information-cost
trade-off, i.e.\ the residual label information it supplies
increases with resolution and/or hop depth while its estimation
cost rises alongside, yielding an interior optimum. A
complementarity result establishes that no channel can substitute
for another even in the infinite-data limit. ATLAS attains the best 
average rank (5.44) among 17 methods on eight medium-scale datasets,
and matches the top rank (4.40) among scalable baselines at
million-node scale without per-epoch propagation. These results
position ATLAS as a robust, scalable, modular alternative across
heterogeneous structural regimes: selecting complementary structural
channels, rather than propagating through one topology operator,
reduces both training and inference to simple MLP computation.

\end{abstract}



\maketitle

\section{Introduction}
\label{sec:intro}

Node classification is a fundamental problem in graph learning.
Accurate classification draws on two complementary sources, namely, node
attributes and structural context. Multi-Layer Perceptrons (MLPs)
exploit only attributes and are structure-blind. Graph Neural
Networks (GNNs) exploit structure through iterative message
passing, under the assumption that a node shares its neighbors'
class. On heterophilic graphs, where edges need not connect
same-label nodes, this assumption is not merely unhelpful but
actively misleading~\cite{zhu2020beyond,platonov2023characterizing}.
Roman-Empire is one such case, where role-like labels alternate
along the graph rather than clustering.

Existing heterophily-aware architectures typically rewire the graph
or redesign the aggregation operator, including, signed or frequency-adaptive
filters, decoupled ego/neighbor embeddings, global correlation
operators~\cite{bo2021beyond,li2022finding}. Because these
mechanisms still route prediction through the adjacency matrix at
every layer, they cost as much as plain message passing, and some
add further structural computation on top. Sampling-based methods
reduce this cost~\cite{chiang2019cluster,zeng2020graphsaint} but
inherit the same smoothness assumption and its failure mode.
Compounding the problem, edge homophily $h_e$ is not a reliable
guide to whether structure helps. Similar $h_e$ scores can mask
large performance differences, and a low score does not imply the
topology is uninformative~\cite{platonov2023characterizing,
platonov2023critical,zhu2020beyond}.

\textbf{Our Approach.} Rewiring the graph, signing its filters, or learning a
global correlation operator all treat heterophily as a single
defect to be corrected, paying for that correction with repeated
passes over the adjacency matrix. But homophily and heterophily form
a continuum, not a binary, and no single corrective method applied
uniformly can suffice.

\begin{figure*}[!t]
\centering
\includegraphics[width=1\linewidth,height=0.25\textheight,keepaspectratio]{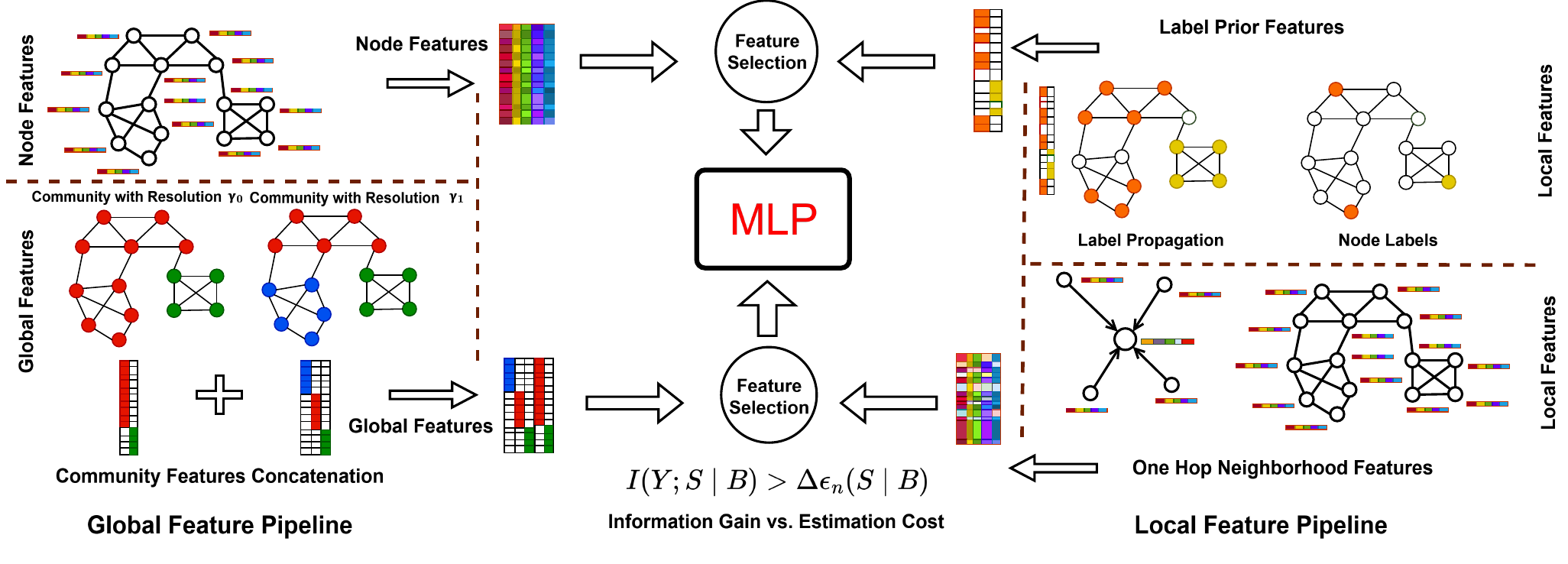}
\caption{Overview of the ATLAS pipeline. ATLAS extracts topology as three precomputed, complementary channels;  multi-resolution community structure (global), one-hop neighbor features, and multi-hop label priors (local). Each channel passes through a feature-selection step governed by the information-versus-estimation-cost. Surviving channels are concatenated with node features and fed into a single MLP for propagation-free training and inference.}
\label{fig:pipeline}
\end{figure*}

To address this, we present ATLAS (Adaptive Topology-based Learning
at Scale) (see Figure~\ref{fig:pipeline}), a scalable, modular
alternative that performs well across heterogeneous structural
regimes without per-epoch propagation. The key idea is to keep the
graph structure unchanged and instead extract several qualitatively
different structural views as candidate \emph{channels}: community
membership, one-hop attribute context, and label evidence. We treat
locality as a free parameter, letting each graph set its own bounds
via an information-versus-cost criterion. Each channel is computed
once during preprocessing -- one hop for NF, $H$ hops for LPF --
giving propagation-free training and adjacency-free inference after
one-time preprocessing, in which the full transductive graph and
training labels are used to build the precomputed features.

\textbf{Choosing the channels.} Community membership captures
mesoscopic structure;\footnote{Separate channels are created per
resolution when multiple community scales carry information.}
one-hop neighbor features (NF) and multi-hop label priors (LPF)
capture information closer to each node. These channels are
structurally complementary. Community features are constant within
a block and so miss any label variation among co-members, while NF
and LPF are built from a node's reachable surroundings and so miss
long-range partition structure.  Estimation cost grows with how much of a
channel is kept, more community resolutions, deeper label-prior
walks, so for each channel we use a cost term to decide how far to
go.

Our main contributions are as follows:

\begin{enumerate}
\item \textbf{An information-cost theory of topology channels}
(Section~2). We show the three channels are informationally
complementary in the worst case, then replace this
idealized picture with a finite-sample inclusion criterion
 that trades information gain against estimation cost.
This criterion motivates
the search procedure below.

\item \textbf{A selection-driven, propagation-free channel
framework} (Section~3). ATLAS treats community resolution and LPF
hop depth as a search space; an adaptive, label-free resolution
search locates informative resolutions, and a 
training-only feature-reduction step prunes the result to
task-aligned dimensions. The output feeds a compact MLP, giving
propagation-free training and adjacency-free inference.

\item \textbf{An empirical evaluation across structural-bias
regimes} (Section~4). We evaluate ATLAS on 19 benchmarks spanning
homophilic, heterophilic, and million-node graphs, showing in most cases it is 
competitive with or exceeding strong message-passing,
heterophily-aware, and scalable decoupled/sampling baselines. 
\end{enumerate}
%

\section{Theoretical Analysis}
\label{sec:theory}

ATLAS converts graph topology into three feature channels:
multi-resolution community membership, one-hop neighbor features
(NF), and multi-hop label priors (LPF), and trains an MLP on the
augmented representation. We use theoretical analysis to answer
three questions: (i) when does including a channel improve
classification, (ii) why are multiple channels and resolutions
required, and (iii) why can no channel be dropped in favor of the
other two. Due to space constraints, most proofs are in
Appendix.

\textbf{Scope of the theory.} 
Throughout this section, "risk" refers to expected cross-entropy (log-loss) risk, matching ATLAS's training objective. We state this because mutual information has an exact identity with log-loss risk reduction but only an inexact relationship with 0-1 classification risk; results stated via $I(Y;S\mid B)$
describe the training objective, not guarantees about test-set accuracy, which we validate empirically instead (Section~4).

\begin{table}[t]
\centering
\footnotesize
\setlength{\tabcolsep}{4pt}
\renewcommand{\arraystretch}{1.1}

\caption{Notation used throughout the paper.}
\label{tab:notation}

\begin{tabularx}{\columnwidth}{
>{\raggedright\arraybackslash}p{2.6cm}
>{\raggedright\arraybackslash}X
}
\toprule
Symbol & Meaning \\
\midrule

$G=(V,E)$, $A$ & Graph and adjacency matrix \\
$X$, $Y$ & Node attributes, node labels \\
$\tilde A = D^{-1}A$ & Row-normalized (random-walk) adjacency \\

$C^{(\gamma)}$, $\mathcal H=\{C^{(1)},\dots,C^{(T)}\}$ 
& Community partition at resolution $\gamma$; retained family \\

$O^{(\gamma)}$ & One-hot encoding of $C^{(\gamma)}$ \\
$W^{(\gamma)}$, $E^{(\gamma)}$ & Learned projection and embedding \\

$X_{\mathrm{NF}} = \tilde A X$ 
& One-hop neighbor-feature channel \\

$H_{\max}$, $Y^{(h)}$, $Y_{\mathrm{LPF}}$ 
& LPF hop bound, $h$-th diffusion step, stacked LPF channel \\

$H(\cdot)$, $I(\cdot;\cdot)$ & Entropy and mutual information \\


$\rho$ 
& Same-class neighbor fraction (Lemma~4, Theorem~4) \\

$\tau$ & ATLAS-FR retained-dimension fraction \\


$L_{\mathrm{ff}}$, $d_{\mathrm{hid}}$ 
& MLP layer count and hidden width \\

$Q_{\min}$, $\Delta_{\max}$, $[a,b]$ 
& Adaptive-search hyperparameters (Algorithm~1) \\

\bottomrule
\end{tabularx}
\end{table}

\paragraph{Statement 1.} Given infinite data, retaining every
community resolution and hop of label diffusion does not decrease the
information available to a log-loss-optimal classifier. In practice, with a finite training
set, some estimation error is incurred. A channel should be added
only if the information gained exceeds this error.

\begin{lemma}[Complementarity of information]
\label{lem:1}
$I(Y;B,S) \ge \max\{I(Y;B), I(Y;S)\}$, with $I(Y;B,S) - I(Y;B) =
I(Y;S\mid B)$.
\end{lemma}

\begin{theorem}[Log-loss Bayes risk monotonicity]
\label{thm:1}
Let $R^*(U) := H(Y\mid U)$ denote the expected cross-entropy risk of
the Bayes-optimal (log-loss-minimizing) classifier using
representation $U$. Then $R^*(B,S) \le R^*(B)$ always, and
$R^*(B,S) < R^*(B)$ strictly whenever $I(Y;S\mid B) > 0$: the
ideal log-loss classifier is never harmed by adding channel $S$ to
base $B$, and is strictly helped whenever $I(Y;S\mid B)>0$.
\end{theorem}

\emph{Remark.} 
Theorem~\ref{thm:1} is exact for log-loss because $R^*(U) = H(Y\mid
U)$ is itself an entropy, so the risk reduction $R^*(B) - R^*(B,S)$
equals $I(Y;S\mid B)$ by definition. This identity does
\textbf{not} extend to 0-1 classification risk $R^*_{01}(U) =
\mathbb E[1-\max_y P(Y=y\mid U)]$. Positive conditional mutual
information means the conditional label distribution changes, but
not necessarily which class is most likely, so $I(Y;S\mid B)>0$ need
not imply $R^*_{01}(B,S) < R^*_{01}(B)$. The log-loss theory below
and the classification-accuracy results of Section~\ref{exp} are connected
empirically, in a practical sense. 

The next theorem replaces the population quantity $I(Y;B,S) -
I(Y;B)$ with its finite-sample analogue, still under log-loss.

\begin{theorem}[Finite-sample risk decomposition]
\label{thm:2}
\[
\mathbb E_S\!\left[R_{B,S}(\hat f_{B,S}) - R_B(\hat f_B)\right]
= -I(Y;S\mid B) + \Delta\epsilon_n(S\mid B).
\]
Here $R_B(\hat f_B)$ is the expected cross-entropy test risk of
$\hat f_B$, trained on base features $B$ alone, and
$R_{B,S}(\hat f_{B,S})$ is the same quantity for $B$ and channel $S$
jointly. $\Delta\epsilon_n(S\mid B) := \epsilon_n(B,S) -
\epsilon_n(B)$ is the  difference in excess risk
(estimation error). Including $S$ reduces expected test risk iff
\[
I(Y;S\mid B) > \Delta\epsilon_n(S\mid B).
\]
\end{theorem}

\emph{Remark (scope of Theorem~\ref{thm:2}).} Theorem~\ref{thm:2} is
a \textbf{decomposition identity}, not an independently
characterized bound: $\Delta\epsilon_n(S\mid B)$ is \emph{defined}
as the gap between excess risks, with no closed form given here. Its
value is in isolating \emph{which} quantities govern the inclusion
decision, estimation cost and motivating the
heuristics of  ATLAS's adaptive resolution search.

\paragraph{Statement 2.}  Every time community resolution increases,
the partition can only become more informative (Lemma~\ref{lem:2}).
But each new block adds a one-hot input dimension the MLP must learn
weights for, so entropy -- and with it, estimation cost -- increases
(Lemma~\ref{lem:3}) toward a bound set by the number of nodes.
Information saturates at $H(Y)$; entropy does not, until every node
is its own community; somewhere in between lies the best resolution.

\begin{lemma}[Information monotonicity under refinement]
\label{lem:2}
For $C' \preceq C$ and any $Z$: $I(Y;C'\mid Z) \ge I(Y;C\mid Z)$.
\end{lemma}

\begin{lemma}[Entropy monotonicity under refinement, bounded]
\label{lem:3}
If $C' \preceq C$, then $H(C') \ge H(C)$, strictly for any strict
split. For a graph on $n$ nodes, $H(C) \le \log n$ for every
partition $C$, with equality approached as $C$ tends to the
all-singletons partition.
\end{lemma}

\begin{theorem}[NMI refinement condition]
\label{thm:3}
For $C' \preceq C$, $\Delta I = I(C';Y) - I(C;Y)$, $\Delta H = H(C')
- H(C) > 0$:
\[
\mathrm{NMI}(C',Y) > \mathrm{NMI}(C,Y) \iff \frac{\Delta I}{\Delta H}
> \frac{\mathrm{NMI}(C,Y)}{2}.
\]
Because $I(C;Y)$ saturates at $H(Y)$ while $H(C)$ increases toward
its finite ceiling $\log n$ (Lemma~\ref{lem:3}) as resolution grows,
Theorem~\ref{thm:3}'s condition eventually fails for any resolution
fine enough: coarse partitions waste available signal, over-refined
ones spend estimation budget on near-duplicate blocks as $H(C)$
approaches $\log n$.
\end{theorem}

\emph{Remark (idealized nesting).}
Lemma~\ref{lem:2}--\ref{lem:3} and Theorem~\ref{thm:3} are proved
under the idealized assumption of \emph{nested} refinement, $C'
\preceq C$. In reality, partitions across
resolutions are not guaranteed to nest. 

\paragraph{Statement 3.} NF is one hop, so estimation cost is fixed. Under a specific probabilistic model of neighbor attributes, NF helps to the extent that a node's same-class neighbor fraction is high enough for denoising to outweigh noise from different-class neighbors.

\begin{lemma}[Variance reduction under one-hop averaging]
Let a single class-match indicator $M \sim \mathrm{Bernoulli}(\rho)$ govern
node $i$'s neighborhood: when $M = 1$, $X_j = \mu_{Y_i} + \eta_j$ for
\emph{every} neighbor $j \in N(i)$; when $M = 0$, $X_j = \mu_{Y_j} + \eta_j$
for every neighbor, with $\mu_{Y_j} - \mu_{Y_i} =: \Delta$ a single mismatch
offset shared across the neighborhood, of mean zero and variance $v$. Here
$\eta_j$ are i.i.d.\ zero-mean noise of variance $\sigma^2$, independent of
$M$ and $\Delta$. For node $i$ with degree $d_i$:
\[
\mathrm{Var}(X_{\mathrm{NF},i} \mid Y_i)
= \rho^2\,\frac{\sigma^2}{d_i} + (1-\rho)\,v.
\]
\end{lemma}

\noindent\textbf{Consistency check.} The formula behaves sensibly at the
extremes; at $\rho = 1$ it reduces to $\sigma^2/d_i$. At $\rho = 0$ it reduces to $v$, \emph{independent of $d_i$}:
with no matched neighbors, the entire neighborhood shares the same offset
$\Delta$, so averaging over $d_i$ neighbors provides no denoising of the
mismatch term. 

\begin{theorem}[Population benefit condition for NF]
Assume in addition that, conditional on $Y_i$, the class-conditional
distributions of $X_i$ and $X_{\mathrm{NF},i}$ are well approximated by
Gaussians whose conditional mutual information with $Y_i$ is determined by
their within-class variance (a second-moment / Gaussian-channel
approximation). Under this assumption, writing
\[
g(\rho) := \mathrm{Var}(X_{\mathrm{NF},i} \mid Y_i)
= \rho^2\,\frac{\sigma^2}{d_i} + (1-\rho)\,v
\]
from Lemma~4:
\[
I(Y; X_{\mathrm{NF}} \mid X) \approx \max\left\{0,\ \tfrac{1}{2}
\log \frac{\sigma^2}{g(\rho)}\right\},
\]
which is non-negative by construction. This is strictly positive and
strictly increasing in $\rho$ once $\rho$ exceeds the crossover point
$\rho^*$ --- the unique root on $(0,1)$, when it exists, of $g(\rho) =
\sigma^2$ --- and is exactly zero for $\rho \le \rho^*$, including the
limit $\rho \to 0$ whenever $v \ge \sigma^2$.
\end{theorem}

\emph{Remark.} The $\max\{0,\cdot\}$ floor is necessary. Without it, the log-ratio approximation can return a
negative value when $g(\rho) > \sigma^2$. With the floor, the $\rho\to 0$ limit
follows directly from the formula. 
Variance reduction (Lemma 4) is necessary
for $X_{\mathrm{NF}}$ to help, and a crossover in $\rho$ is expected
on general grounds, though the exact threshold $\rho^*$ relies on
the stated approximation. 

\paragraph{Statement 4.}  In the ideal case, a longer random walk only
adds information (Lemma~\ref{lem:5}). In practice,  the walk converges to a
node-independent vector at a rate set by the graph's spectral gap,
so -- under a Lipschitz assumption on the downstream readout -- the
achievable risk reduction from that information shrinks
exponentially with depth (Theorem~\ref{thm:5}).

\begin{lemma}[Information monotonicity in hop depth]
\label{lem:5}
For every $H\ge 1$: $I\big(Y; Y^{(1)},\dots,Y^{(H+1)} \mid X\big) \ge
I\big(Y; Y^{(1)},\dots,Y^{(H)} \mid X\big)$.
\end{lemma}

\begin{theorem}[Geometric oversmoothing] 
\label{thm:5}
For connected, non-bipartite $G$ with second-largest eigenvalue modulus
$\lambda$ of $\tilde{A}$, $Y^{(h)}$ converges to a node-independent vector
at rate $\lambda^h$.

\end{theorem}

\emph{Remark (directed graphs).} Theorem~\ref{thm:5}'s $\lambda^h$
rate is derived for undirected graphs, where $\tilde A$ is
diagonalizable with real spectrum via similarity to a symmetric
matrix. LPF's directional extension $\mathcal T_\alpha$ (Section~3)
uses $\tilde A_{\mathrm{in}}$ and $\tilde A_{\mathrm{out}}$, which
are generally non-normal and their powers can exhibit transient,
non-monotonic behavior. The strong
empirical results on Arxiv-Year and Snap-Patents (Table~\ref{tab:large_scale_heterophily}) provide empirical validation.

Statements 2--4 are specific cases of Theorem~\ref{thm:2} applied to
different channels, with the \emph{information} term instantiated
differently in each: for communities, information saturates while
the cost term (entropy) grows toward its ceiling; for NF, both
information and cost are fixed regardless of resolution; for LPF, the \emph{cost} term
grows only linearly with hop depth, while it is the
\textbf{information} term that decays geometrically under
oversmoothing.  Because the channels sit at different points on
the homophily--heterophily spectrum, a channel can have a favorable
information/cost trade-off on one graph and an unfavorable one on
another.

\paragraph{Statement 5.} Community, one-hop features, and label
priors are structurally complementary in the sense that each can
encode information the other two cannot recover in the worst case,
so none is \emph{universally} redundant. However,   all three may not be simultaneously useful on any
particular graph.

\begin{theorem}[Structural complementarity]
\label{thm:6}
Let the multi-resolution communities be denoted by $\mathcal C =
(C^{(\gamma_1)},\dots,C^{(\gamma_T)})$, the one-hop neighbor feature
by $X_{\mathrm{NF}}$, and label prior by $Y_{\mathrm{LPF}} =
(Y^{(1)},\dots,Y^{(H_{\max})})$. There exist joint distributions
$P_1,P_2,P_3$ over $(G,X,Y)$ such that
\begin{align}
I_{P_1}\big(Y;\mathcal C \mid X, X_{\mathrm{NF}}, Y_{\mathrm{LPF}}\big) &> 0,
\tag{a}\\
I_{P_2}\big(Y;X_{\mathrm{NF}} \mid X, \mathcal C, Y_{\mathrm{LPF}}\big) &> 0,
\tag{b}\\
I_{P_3}\big(Y;Y_{\mathrm{LPF}} \mid X, \mathcal C, X_{\mathrm{NF}}\big) &> 0,
\tag{c}
\end{align}
while under each $P_k$ the two channels other than the one being
tested carry zero information about $Y$ beyond $X$ alone.
Consequently no channel is, in general, a deterministic function of,
or informationally subsumed by, the union of the remaining two, even
as $n\to\infty$. Moreover, there exists a distribution $P_4$ under
which all three channels together strictly dominate any two.
\end{theorem}

\emph{Remark (scope).} Theorem~\ref{thm:6} establishes
\emph{existence} of graphs on which each channel is indispensable, but
it does not by
itself justify keeping all three channels on any specific dataset.
That question is answered empirically by the per-channel ablations
(ATLAS vs.\ ATLAS-NF vs.\ ATLAS-LPF vs.\ ATLAS-NF-LPF), which are
the actual evidence for complementarity on real graphs.

\section{Methodology}
\label{sec:methodology}

The theory in Section~\ref{sec:theory} provided an idealized selection criterion. In this section we apply a modularity-based heuristic for adaptive search which is motivated by this criterion.
We use modularity rather than NMI because it can be computed without
labels. Scoring many candidate resolutions against labels (as NMI
would require) risks selection bias and is unreliable when labels
are sparse; modularity avoids both problems since it only needs
graph structure.



\paragraph{\bf Multi-Resolution Community Extraction.} The scale at which topology correlates with labels is unknown a-priori and varies with the type of graph. 
We use {\em multi-resolution communities to identify the correct topological scale.}

Section~\ref{sec:theory} analyzes community refinement under an idealized assumption that refined communities nest perfectly inside coarser ones. In practice, community detection does not guarantee this. We run Louvain~\footnote{Results with other partitioning methods such as Leiden and Metis are given in the Appendix Table ~\ref{tab:detector}} at multiple resolution values $\gamma$. Increasing $\gamma$ typically yields finer partitions, although they may not be nested within the coarser ones. Homophily also varies across datasets, therefore refinement does not always increase label homogeneity. ATLAS runs an adaptive resolution search during preprocessing that identifies a small set of informative resolutions, approximating the idealized refinement structure(s) at a fraction of the cost.

{\em Notation.}
For resolution $\gamma$, Louvain returns a node-to-community assignment $c^{(\gamma)}$ and modularity score $Q^{(\gamma)}$, which measures within-community connectivity strength relative to a random-edge null model. For $\gamma_1 < \gamma_2$, the modularity gap is $\Delta Q(\gamma_1, \gamma_2) = \left| Q^{(\gamma_2)} - Q^{(\gamma_1)} \right|$.
Appendix~\ref{app:community_defs} gives formal definitions.

{\em Adaptive Resolution Search.}
We start the search from $\gamma \in \{0.5, 1.0\}$ (we may include smaller values for graphs with informative low-resolution structure) and refine this set using three hyperparameters; a maximum modularity-gap threshold $\Delta_{\max}$, a minimum modularity threshold $Q_{\min}$, and a target modularity-decrease range $[a, b]$. At each iteration, neighboring resolution pairs are examined: if a gap exceeds $\Delta_{\max}$, the algorithm interpolates by inserting their midpoint; otherwise it extrapolates beyond the current maximum $\gamma$, choosing a step that reduces modularity by an amount sampled from $[a, b]$. The search stops once modularity at the largest $\gamma$ drops below $Q_{\min}$, or no new resolutions are generated. The output is a resolution set $\Gamma$ and community assignments $\{c^{(\gamma)}\}_{\gamma \in \Gamma}$, used as multi-resolution community features for augmentation. The pseudocode  is provided in the Appendix Algorithm~\ref{alg:adaptive-res}.

{\em Feature Augmentation.}
 Community detection yields assignments $c^{(\gamma)} \in \{1, \dots, k_\gamma\}^n$ (Eq.~\ref{eq:community-detect}), one-hot encoded as $O^{(\gamma)}$ (Eq.~\ref{eq:onehot}) and projected to a dense space via trainable $W^{(\gamma)}$ (Eq.~\ref{eq:project}). Concatenating across resolutions gives $E$ (Eq.~\ref{eq:concat}), which is then concatenated with node features $X$ to form $Z_{\mathrm{ATLAS}}$ (Eq.~\ref{eq:zatlas}). An MLP $f_\theta$ outputs logits, mapped to probabilities $\hat{Y}$ via $\phi$ (softmax or sigmoid, Eq.~\ref{eq:yhat}):
\begin{align}
c^{(\gamma)} &= \mathrm{DetectCommunity}(G, \gamma) \label{eq:community-detect} \\
O^{(\gamma)} &= \mathrm{OneHot}(c^{(\gamma)}) \label{eq:onehot} \\
E^{(\gamma)} &= O^{(\gamma)} W^{(\gamma)} \label{eq:project} \\
E &= \big\Vert_{t=1}^{T}E^{(\gamma_t)} \in \mathbb{R}^{n\times (T d_c)} \label{eq:concat} \\
Z_{\mathrm{ATLAS}} &= [X \,\|\, E] \label{eq:zatlas} \\
\hat{Y} &= \phi\big(f_\theta(Z_{\mathrm{ATLAS}})\big) \label{eq:yhat}
\end{align}

\paragraph{Complexity.}
ATLAS incurs a one-time community-detection cost, $O(T \|A\|_0)$ for $T$ resolutions; training and inference follow a MLP over features of dimension $D + T d_c$ with propagation-free training, adjacency-free inference, with cost scaling linearly in $T$ (Table~\ref{tab:complexity}).

\paragraph {\bf Neighbor Features and Label Priors.} Neighbor features and label priors exist to recover that within-block signal that community cannot identify. Both use the row-normalized operator $\tilde{A} = \mathrm{RowNorm}(A) = D^{-1}A$, concatenated with $Z_{\mathrm{ATLAS}}$.

{\em Neighbor Features (ATLAS-NF)}
A single hop of neighbor attributes, $X_{\mathrm{NF}}[i] = \mathbb{E}[X_{X_1} \mid X_0 = i]$, supplies per-node local context.


\begin{align}
X_{\mathrm{NF}} &= \tilde{A} X, \qquad
Z_{\mathrm{ATLAS\text{-}NF}} = [Z_{\mathrm{ATLAS}} \,\|\, X_{\mathrm{NF}}] \label{eq:atlas-nf}
\end{align}

{\em Label Priors (ATLAS-LPF)}
Masked training labels diffuse over increasing hop depth. Since $\tilde{A}$ is a stochastic transition matrix, hop $h$ corresponds to an $h$-step random walk, $Y^{(h)}[i] = \mathbb{E}[Y^{(0)}[X_h] \mid X_0 = i]$ and labels from every hop are kept as a separate feature:

\begin{align}
Y^{(0)} &= \mathrm{diag}(m_{\mathrm{tr}})\, Y \label{eq:y0} \\
Y^{(h)} &= \tilde{A}\, Y^{(h-1)}, \quad h = 1, \dots, H \label{eq:yh} \\
Y_{\mathrm{LPF}}&=\big\Vert_{h=1}^{H}\mathbf{Y}^{(h)} \in \mathbb{R}^{n\times (HC)} \label{eq:ylpf}\\
Z_{\mathrm{ATLAS\text{-}LPF}} &= [Z_{\mathrm{ATLAS}} \,\|\, Y_{\mathrm{LPF}}] \label{eq:zatlaslpf}
\end{align}

Only training labels enter $Y^{(0)}$, so LPF is a legitimate semi-supervised feature. Stacking both channels gives $Z_{\mathrm{ATLAS\text{-}NF\text{-}LPF}} = [Z_{\mathrm{ATLAS}} \,\|\, X_{\mathrm{NF}} \,\|\, Y_{\mathrm{LPF}}]$.

\paragraph{Directed graphs.}
For directed graphs, edge orientation may contain useful signals. An optional directional extension separates incoming and outgoing neighborhoods via a learned gate,
\[
\mathcal{T}_\alpha(\cdot) = \alpha \tilde{A}_{\mathrm{in}}(\cdot) + (1-\alpha)\tilde{A}_{\mathrm{out}}(\cdot), \qquad \alpha = \sigma(\tilde{\alpha}),
\]
and the resulting directional NF/LPF features are concatenated with the standard representation. 

\paragraph{\bf Feature Reduction.} ATLAS-FR removes redundant augmentation dimensions arising from over-refined community structure or dense neighborhood signals. Using training nodes only, we compute ANOVA $F$-scores and retain the top $\tau$ fraction of dimensions, selected independently for the community and NF features. 

\paragraph{\bf Design Rationale.} ATLAS separately concatenates the three channels, instead of combining them because each captures signal the others structurally cannot. Community features capture mesoscale structure but are blind to within-block variation; neighbor attributes ($X_{\mathrm{NF}} = \tilde{A}X$) recover that within-block signal by denoising each node under attribute homophily; label priors ($Y^{(h)} = \tilde{A}^h Y^{(0)}$) supply class-compatibility signal independent of homophily. Exposing these as separate, explicit MLP inputs lets us diagnose which channel carries predictive signal on a given graph. 
\section{Empirical Evaluation}
\label{exp}


{\bf Structural bias.}
We define structural bias as the degree to which community structure remains informative for prediction as partitions are refined. This is measured by sweeping the minimum modularity threshold $Q_{\min}$ in the adaptive search (lower $Q_{\min}$ retains more, finer resolutions). Graphs exhibit \emph{high} structural bias when refinement improves performance, \emph{low} structural bias when gains plateau, and \emph{negative} structural bias when refinement degrades performance or leaves it unchanged. More details in Section~\ref{refinement}.


\textbf{Datasets.} We evaluate on 19 benchmarks in total: 8 medium-scale homophilic/heterophilic graphs, 5 million-node graphs, and 6 large non-homophilous graphs. We report ROC-AUC for the binary datasets (Tolokers, Questions), micro-F1 for Flickr, Reddit, Yelp, AmazonProducts, and Accuracy elsewhere, following standard protocols from prior work. The medium-scale group covers Cora, Actor, Questions, Chameleon-Filtered, Squirrel-Filtered, Amazon-Ratings, Tolokers, and Roman-Empire, using splits from Platonov et al.~\cite{platonov2023critical} where applicable and a 60/20/20 split otherwise; the large-scale group covers Flickr, Reddit, Yelp, AmazonProducts, and ogbn-products, using GraphSAINT/GraphSAGE/OGB splits~\cite{hamilton2017inductive,hu2020ogb,zeng2020graphsaint}; the non-homophilous group covers the six LINKX benchmarks~\cite{lim2021linkx} (Penn94, Pokec, Arxiv-Year, Snap-Patents, Genius, Twitch-Gamers). Full statistics are given in Appendix Tables~\ref{tab:exp-dataset-homophily} -~\ref{tab:exp-dataset-linkx}.

\textbf{Baselines.} Baselines are grouped by modeling regime and aligned to each research question. For Q1, we compare against homophily-oriented GNNs (GCN, GraphSAGE, GAT) and heterophily-oriented models (H2GCN, LINKX, GPR-GNN, FSGNN, GloGNN, FAGCN, GBK-GNN, JacobiConv, ACM-GCN, OrderedGNN, M2M-GNN, CMGNN). For Q2, we compare MLP, GCN, GAT, GPR-GNN, ACM-GCN, and LINKX against ATLAS and its NF/LPF/directional variants. For Q3, we compare against precomputation-based decoupled methods (SGC, SIGN, SAGN, GAMLP) and sampling-based methods (GraphSAGE, ClusterGCN, GraphSAINT, LABOR, GRAPES). Full descriptions of all baselines are given in Appendix ~\ref{sec:homophily-models}--~\ref{sec:decoupling-methods}.

All MLP heads use $L$ layers of width $d_{\mathrm{hid}}$ with dropout $p$: each of the first $L-1$ layers applies Linear $\to$ LayerNorm $\to$ GELU $\to$ Dropout, and the final layer is a linear classifier over $C$ classes.


\begin{table*}[!t]
\centering
\scriptsize
\setlength{\tabcolsep}{3.0pt}          
\renewcommand{\arraystretch}{1.08}     
\caption{Eight-benchmark comparison across homophily regimes.
Rightmost column reports the average rank across the eight datasets (lower is better).
Cells highlighted in \cellcolor{yellow!20}{yellow} indicate the best score for each dataset (and best Avg. Rank);
\cellcolor{secondbest}{blue} indicates the second-best. Blank entries indicate no improvement by community or neighborhood feature reduction.
}
\label{tab:results_transposed_full}
\vspace{-1mm}

\scalebox{0.95}{%
\begin{tabular}{l c c c c c c c c c}
\toprule
& \multicolumn{2}{c}{\cellcolor{colA}\textbf{High structural bias}}
& \multicolumn{4}{c}{\cellcolor{colB}\textbf{Low structural bias}}
& \multicolumn{2}{c}{\cellcolor{colC}\textbf{Negative structural bias}}
& \multicolumn{1}{c}{\textbf{Overall}} \\
\cmidrule(lr){2-3}\cmidrule(lr){4-7}\cmidrule(lr){8-9}\cmidrule(lr){10-10}

Model
& \cellcolor{colA}\shortstack{Cora \\ \scriptsize $h_e=0.810$}
& \cellcolor{colA}\shortstack{Tolokers \\ \scriptsize $h_e=0.595$}
& \cellcolor{colB}\shortstack{Chameleon (Filt.) \\ \scriptsize $h_e=0.236$}
& \cellcolor{colB}\shortstack{Amazon-Ratings \\ \scriptsize $h_e=0.380$}
& \cellcolor{colB}\shortstack{Questions \\ \scriptsize $h_e=0.84$}
& \cellcolor{colB}\shortstack{Squirrel (Filt.) \\ \scriptsize $h_e=0.207$}
& \cellcolor{colC}\shortstack{Actor \\ \scriptsize $h_e=0.216$}
& \cellcolor{colC}\shortstack{Roman-Empire \\ \scriptsize $h_e=0.047$}
& \shortstack{Avg. Rank \\ \scriptsize $\downarrow$} \\
\midrule

MLP
& 75.44 $\pm$ 1.97 & 72.97 $\pm$ 0.90
& 36.00 $\pm$ 4.69 & 39.83 $\pm$ 0.48
& 71.87 $\pm$ 0.70 & 35.65 $\pm$ 2.38
& 38.48 $\pm$ 0.93 & 66.22 $\pm$ 0.53
& 14.75 \\

\midrule

GCN
& 87.01 $\pm$ 1.04 & 74.93 $\pm$ 1.32
& 37.11 $\pm$ 3.04 & 42.78 $\pm$ 0.14
& 76.09 $\pm$ 1.27 & 32.70 $\pm$ 1.73
& 28.49 $\pm$ 0.91 & 45.68 $\pm$ 0.38
& 16.25 \\

SAGE
& 87.50 $\pm$ 0.87 & 80.95 $\pm$ 0.92
& 38.83 $\pm$ 4.26 & 44.67 $\pm$ 0.51
& 76.44 $\pm$ 0.62 & 33.32 $\pm$ 1.75
& 34.08 $\pm$ 1.07 & 76.21 $\pm$ 0.65
& 12.13 \\

GAT
& 87.74 $\pm$ 0.88 & 75.31 $\pm$ 1.35
& 37.18 $\pm$ 3.44 & 43.25 $\pm$ 0.85
& 77.43 $\pm$ 1.20 & 32.61 $\pm$ 2.06
& 29.11 $\pm$ 1.23 & 47.16 $\pm$ 0.66
& 14.50 \\

\midrule

H2GCN
& 87.52 $\pm$ 0.61 & 73.35 $\pm$ 1.01
& 26.75 $\pm$ 3.64 & 36.47 $\pm$ 0.23
& 63.59 $\pm$ 1.46 & 35.10 $\pm$ 1.15
& 38.85 $\pm$ 1.17 & 60.11 $\pm$ 0.52
& 15.00 \\

LINKX
& 82.62 $\pm$ 1.44 & 81.15 $\pm$ 1.23
& 42.34 $\pm$ 4.13 & 52.66 $\pm$ 0.64
& 71.96 $\pm$ 2.07 & 40.10 $\pm$ 2.21
& 35.64 $\pm$ 1.36 & 56.15 $\pm$ 0.93
& 10.25 \\

GPR-GNN
& 79.51 $\pm$ 0.36 & 72.94 $\pm$ 0.97
& 39.93 $\pm$ 3.30 & 44.88 $\pm$ 0.34
& 55.48 $\pm$ 0.91 & 38.95 $\pm$ 1.99
& \cellcolor{secondbest}39.30 $\pm$ 0.27 & 64.85 $\pm$ 0.27
& 13.25 \\

FSGNN
& 87.51 $\pm$ 1.21 & \cellcolor{secondbest}82.76 $\pm$ 0.61
& 40.61 $\pm$ 2.97 & 52.74 $\pm$ 0.83
& \cellcolor{yellow!20}\textbf{78.86 $\pm$ 0.92} & 35.92 $\pm$ 1.32
& 37.65 $\pm$ 0.79 & 79.92 $\pm$ 0.56
& 6.75 \\

GloGNN
& 88.31 $\pm$ 1.13 & 73.39 $\pm$ 1.17
& 25.90 $\pm$ 3.58 & 36.89 $\pm$ 0.14
& 65.74 $\pm$ 1.19 & 35.11 $\pm$ 1.24
& 37.70 $\pm$ 1.40  & 59.63 $\pm$ 0.69
& 15.00 \\

FAGCN
& 88.85 $\pm$ 1.36 & 77.75 $\pm$ 1.05
& 41.90 $\pm$ 2.72 & 44.12 $\pm$ 0.30
& 77.24 $\pm$ 1.26 & \cellcolor{secondbest}41.08 $\pm$ 2.27
& 31.59 $\pm$ 1.37 & 65.22 $\pm$ 0.56
& 9.13 \\

GBK-GNN
& 87.09 $\pm$ 1.52 & 81.01 $\pm$ 0.67
& 39.61 $\pm$ 2.60 & 45.98 $\pm$ 0.71
& 74.47 $\pm$ 0.86 & 35.51 $\pm$ 1.65
& 38.47 $\pm$ 1.53 & 74.57 $\pm$ 0.47
& 10.38 \\

JacobiConv
& \cellcolor{secondbest}89.61 $\pm$ 0.96 & 68.66 $\pm$ 0.65
& 39.00 $\pm$ 4.20 & 43.55 $\pm$ 0.48
& 73.88 $\pm$ 1.16 & 29.71 $\pm$ 1.66
& \cellcolor{yellow!20}\textbf{41.17 $\pm$ 0.64} & 71.14 $\pm$ 0.42
& 11.50 \\

ACM-GCN
& 87.91 $\pm$ 0.95 & 74.95 $\pm$ 1.16
& \cellcolor{secondbest}42.73 $\pm$ 3.59 & 52.49 $\pm$ 0.24
& 62.91 $\pm$ 2.10 & 39.79 $\pm$ 2.15
& 36.12 $\pm$ 1.09 & 71.89 $\pm$ 0.61
& 9.75 \\

OrderedGNN
& 88.37 $\pm$ 0.75 & 75.60 $\pm$ 1.36
& 41.51 $\pm$ 4.15 & 51.15 $\pm$ 0.46
& 75.09 $\pm$ 1.04 & 36.94 $\pm$ 2.94
& 37.99 $\pm$ 1.01 & 82.88 $\pm$ 0.71
& 8.31 \\

M2M-GNN
& 88.12 $\pm$ 1.02 & 80.85 $\pm$ 0.08
& 41.73 $\pm$ 3.08 & 51.58 $\pm$ 0.55
& 71.27 $\pm$ 1.41 & 37.38 $\pm$ 1.14
& 36.72 $\pm$ 1.60 & \cellcolor{secondbest}84.17 $\pm$ 0.49
& 9.13 \\

CMGNN
& 85.76 $\pm$ 1.09 & 76.38 $\pm$ 2.42
& 42.15 $\pm$ 4.84 & 52.57 $\pm$ 0.82
& 56.17 $\pm$ 1.09 & 40.25 $\pm$ 1.63
& 36.82 $\pm$ 0.78 & \cellcolor{yellow!20}\textbf{84.42 $\pm$ 1.66}
& 9.13 \\

\midrule

\textbf{ATLAS}
& 88.37 $\pm$ 0.63 & 82.19 $\pm$ 0.73
& \cellcolor{yellow!20}\textbf{42.87 $\pm$ 3.55} & \cellcolor{yellow!20}\textbf{53.17 $\pm$ 0.81}
& 73.26 $\pm$ 0.83 & \cellcolor{yellow!20}\textbf{41.24 $\pm$ 1.49}
& 38.48 $\pm$ 0.93 & 66.22 $\pm$ 0.53
& \cellcolor{yellow!20}\textbf{5.44} \\

\textbf{ATLAS-NF}
& 89.08 $\pm$ 0.44 & \cellcolor{yellow!20}\textbf{82.88 $\pm$ 0.78}
& 40.02 $\pm$ 2.79 & 52.30 $\pm$ 0.64
& \cellcolor{secondbest}78.17 $\pm$ 1.12 &  38.13 $\pm$ 1.16
& 34.26 $\pm$ 0.97 & 78.10 $\pm$ 0.59
& 6.88 \\

\textbf{ATLAS-LPF}
& 88.46 $\pm$ 0.49 & 82.30 $\pm$ 0.76
& 42.69 $\pm$ 4.07 & \cellcolor{secondbest}52.81 $\pm$ 0.37
& 74.13 $\pm$ 0.92 & 40.52 $\pm$ 2.10
& 38.44 $\pm$ 1.10 & 70.22 $\pm$ 0.37
& \cellcolor{secondbest}5.63 \\

\textbf{ATLAS-LPF-NF}
& 89.20 $\pm$ 0.39 & 82.65 $\pm$ 0.92
& 40.23 $\pm$ 3.78 & 51.91 $\pm$ 0.46
& 77.35 $\pm$ 0.83 & 38.33 $\pm$ 1.53
& 34.72 $\pm$ 1.53 & 79.04 $\pm$ 0.72
& 6.88 \\
\textbf{$\text{ATLAS}$-{FR}}

& \cellcolor{yellow!20}\textbf{90.31 $\pm$ 0.79}& --
& -- & --
& -- & --
& -- & 80.07 $\pm$ 0.51
& \\
\midrule

ATLAS--MLP
& +12.93 & +9.22 & +6.87 & +13.34 & +1.39 & +5.59 & +0.00 & +0.00
& \textemdash \\

ATLAS--GCN
& +1.36 & +7.26 & +5.76 & +10.39 & -2.83 & +8.54 & +9.99 & +20.54
& \textemdash \\

\bottomrule
\end{tabular}%
} 

\vspace{-1mm}
\end{table*}

\paragraph {\bf Q1: Accuracy Across Structural-Bias Regimes.} Table~\ref{tab:results_transposed_full} reports results across the eight medium-scale benchmarks, grouped into high, low, and negative structural-bias. The ATLAS variants double as a diagnostic toolbox. MLP isolates attribute signal alone; ATLAS adds multi-resolution community features; ATLAS-NF and ATLAS-LPF each add one local channel; ATLAS-LPF-NF combines both; and ATLAS-FR applies the training-only feature reduction.

 On high-bias datasets, community features clear the inclusion criterion and ATLAS improves over both MLP and GCN. On low-bias datasets the signal is weaker but still positive, and ATLAS remains competitive by capturing coarse structure. On negative-bias datasets community structure fails the criterion,
so ATLAS's adaptive search discards it and the model closely
tracks MLP accuracy, with any further gains coming from the local
channels. 


{\em ATLAS attains the lowest average rank}(5.44) overall, leading the
low-bias group at 52.64\% average accuracy (+0.61 over FSGNN) while
ATLAS-NF leads the high-bias group at 85.98\% (+0.84 over FSGNN).
ATLAS does not win on every dataset: on Roman-Empire, CMGNN (84.42),
M2M-GNN (84.17), and OrderedGNN (82.88) all exceed even ATLAS-FR's
80.07; on Actor, JacobiConv (41.17), GPR-GNN (39.30), and H2GCN
(38.85) exceed ATLAS's 38.48; and on Questions, FSGNN (78.86)
exceeds ATLAS-NF's 78.17. 

\begin{table}[!t]
\centering
\scriptsize
\setlength{\tabcolsep}{2.2pt}          
\renewcommand{\arraystretch}{1.05}     
\caption{Large-graph performance. We report micro-F1 (\%) on Flickr/Reddit/Yelp/AmazonProducts and Accuracy (\%) on ogbn-products. Cells highlighted in \cellcolor{yellow!20}{yellow} indicate the best score for each dataset (and best Avg. Rank); \cellcolor{secondbest}{blue} indicates the second-best. Blank entries indicate no improvement by community or neighborhood feature reduction.}
\label{tab:exp-sota}
\vspace{-1mm}

\resizebox{\columnwidth}{!}{%
\begin{tabular}{l c c c c c c}
\toprule
& \multicolumn{2}{c}{\cellcolor{colA}\textbf{High structural bias}}
& \multicolumn{1}{c}{\cellcolor{colB}\textbf{Low structural bias}}
& \multicolumn{2}{c}{\cellcolor{colC}\textbf{Negative structural bias}}
& \multicolumn{1}{c}{\textbf{Overall}} \\
\cmidrule(lr){2-3}\cmidrule(lr){4-4}\cmidrule(lr){5-6}\cmidrule(lr){7-7}

Model
& \cellcolor{colA}\shortstack{Reddit \\ \scriptsize $h_e{=}0.756$}
& \cellcolor{colA}\shortstack{ogbn-products \\ \scriptsize $h_e{=}0.808$}
& \cellcolor{colB}\shortstack{Flickr \\ \scriptsize $h_e{=}0.319$}
& \cellcolor{colC}\shortstack{Yelp \\ \scriptsize $h_e{=}0.809$}
& \cellcolor{colC}\shortstack{AmazonProducts \\ \scriptsize $h_e{=}0.116$}
& \shortstack{Avg. Rank \\ \scriptsize $\downarrow$} \\
\midrule

MLP
& 74.35$\pm$0.16
& 61.06$\pm$0.08
& 47.17$\pm$0.11
& \cellcolor{yellow!20}\textbf{65.46$\pm$0.11}
& \cellcolor{yellow!20}\textbf{82.04$\pm$0.02}
& 9.40 \\
GCN
& 93.30$\pm$0.01 & 75.64$\pm$0.21 & 49.20$\pm$0.30 & 37.80$\pm$0.10 & 28.10$\pm$0.50 & 12.80 \\
\midrule

GraphSAGE
& 95.30$\pm$0.10 & 80.61$\pm$0.16 & 50.10$\pm$1.30 & 63.40$\pm$0.60 & 75.80$\pm$0.20 & 6.80 \\
ClusterGCN
& 95.40$\pm$0.10 & 78.62$\pm$0.61 & 48.10$\pm$0.50 & 60.90$\pm$0.50 & 75.90$\pm$0.80 & 8.60 \\
GraphSAINT
& 96.60$\pm$0.10 & 75.36$\pm$0.34 & 51.10$\pm$0.10
& \cellcolor{secondbest}65.30$\pm$0.30
& \cellcolor{secondbest}81.50$\pm$0.10
& \cellcolor{secondbest}5.40 \\
LABOR
& 96.23$\pm$0.05 & 78.59$\pm$0.34 & 51.67$\pm$0.27 & 61.57$\pm$0.67 & 68.25$\pm$0.10 & 6.60 \\
GRAPES
& 94.30$\pm$0.06 & 71.45$\pm$0.20 & 49.54$\pm$0.67 & 44.57$\pm$0.88 & 62.01$\pm$0.06 & 12.00 \\
\midrule

SGC
& 93.51$\pm$0.04 & 67.48$\pm$0.11 & 50.35$\pm$0.05 & 23.56$\pm$0.02 & 22.62$\pm$0.28 & 13.20 \\
SIGN
& 95.95$\pm$0.02 & 80.52$\pm$0.16 & 51.60$\pm$0.11 & 57.98$\pm$0.12 & 74.24$\pm$0.02 & 6.80 \\
SAGN
& 96.48$\pm$0.03 & 81.21$\pm$0.07 & 50.07$\pm$0.11 & 61.55$\pm$0.40 & 76.82$\pm$1.15 & 5.60 \\
GAMLP
& \cellcolor{secondbest}96.62$\pm$0.03
& \cellcolor{yellow!20}\textbf{83.76$\pm$0.19}
& 52.58$\pm$0.12
& 57.84$\pm$1.54
& 75.99$\pm$0.26
& \cellcolor{yellow!20}\textbf{4.40} \\
\midrule

\textbf{ATLAS}
& 95.74$\pm$0.04 & 79.98$\pm$0.14 & 51.56$\pm$0.14
& \cellcolor{yellow!20}\textbf{65.46$\pm$0.11}
& \cellcolor{yellow!20}\textbf{82.04$\pm$0.02}
& \cellcolor{yellow!20}\textbf{4.40} \\
\textbf{ATLAS-NF}
& 94.10$\pm$0.07 & 75.07$\pm$0.30 & \cellcolor{secondbest}52.72$\pm$0.11 & 57.40$\pm$0.21 & 77.83$\pm$0.10 & 8.20 \\
\textbf{ATLAS-LPF}
& \cellcolor{yellow!20}\textbf{96.65$\pm$0.04} & 80.34 $\pm$ 0.33 & 51.77$\pm$0.16 & 60.20$\pm$1.34 & 56.64$\pm$0.28 & 6.00 \\
\textbf{ATLAS-LPF-NF}
& 94.47$\pm$0.10 & 75.14$\pm$0.30 & \cellcolor{yellow!20}\textbf{52.80$\pm$0.17} & 51.92$\pm$2.07 & 53.70$\pm$0.48 & 9.40 \\
\textbf{$\text{ATLAS}$-{FR}}

& -- & \cellcolor{secondbest}81.44 $\pm$ 0.20 & -- & -- & -- &  \\
\midrule

ATLAS--MLP & +21.39 & +18.92 & +4.39 & +0.00 & +0.00 & \textemdash \\
ATLAS--GCN & +2.44  & +4.34  & +2.36 & +27.66 & +53.94 & \textemdash \\
\bottomrule
\end{tabular}%
} 

\vspace{-3mm}
\end{table}

\begin{figure}[!t]
\centering
\includegraphics[width=1\linewidth,height=0.5\textheight,keepaspectratio]{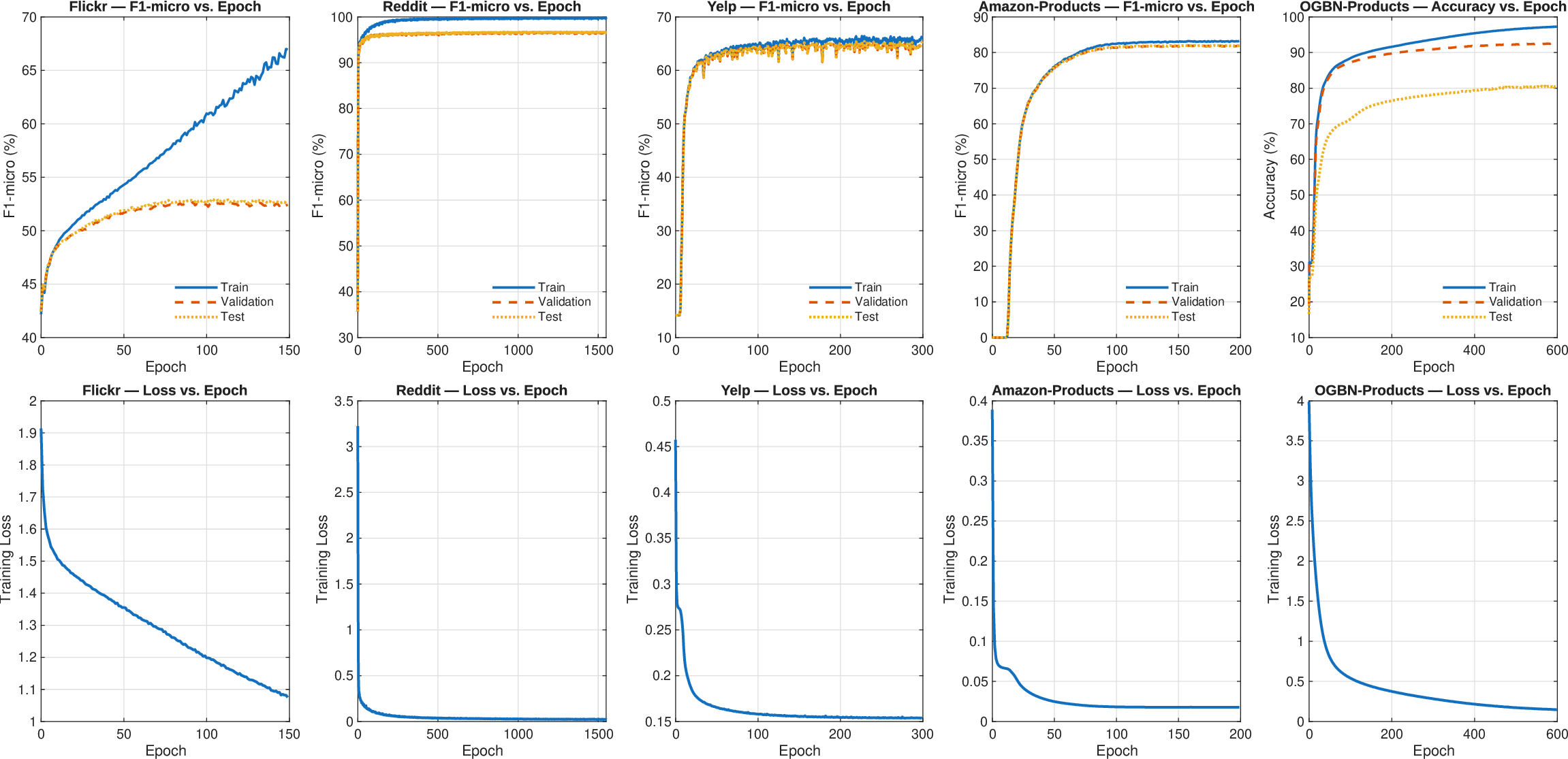}
\caption{The convergence landscape of {ATLAS}.}
\label{fig:convergence-landscape}
\end{figure}

\begin{table}[t]
\centering
\caption{
Complexity comparison.
$N$=\#nodes, $\|A\|_0$=\#edges, $D$=feature dim,
$L$=\#message-passing layers, $L_{ff}$=\#feed-forward layers,
$b$=batch size, 
$K$=\#precomputed hop propagations,
$k$=\#subgraph samples (GraphSAINT),
$T$=\#resolutions, $d_c$=community-embedding dim,
$C$=\#classes, $H$=\#LPF hops.
The two local channels below the rule are optional: each adds a
one-time \textbf{Pre} cost and widens the MLP input, leaving base
\textsc{ATLAS} and its propagation-free training/inference unchanged.}
\label{tab:complexity}
\tiny
\setlength{\tabcolsep}{3pt}
\renewcommand{\arraystretch}{1.05}
\begin{tabular}{@{}lccc@{}}
\toprule
\textbf{Method} & \textbf{Pre} & \textbf{Train/epoch} & \textbf{Mem} \\
\midrule
GCN (full)
& --
& $O\!\left(L(\|A\|_0D+ND^2)\right)$
& $O\!\left(L(ND+D^2)\right)$ \\
ClusterGCN
& $O(\|A\|_0)$
& $O\!\left(L(\|A\|_0D+ND^2)\right)$
& $O\!\left(L(bD+D^2)\right)$ \\
GraphSAINT
& $O(kN)$
& $O\!\left(L(\|A\|_0D+ND^2)\right)$
& $O(LbD)$ \\
\midrule
SAGN
& $O(K\|A\|_0D)$
& $O(L_{ff}NK^2D^2)$
& $O(bL_{ff}KD)$ \\
GAMLP
& $O(K\|A\|_0D)$
& $O(L_{ff}NK^2D^2)$
& $O(bL_{ff}KD)$ \\
\midrule
\textbf{\textsc{ATLAS}}
& $O(T\|A\|_0)$
& $O\!\left(L_{ff}N(D+Td_c)\right)$
& $O\!\left(bL_{ff}(D+Td_c)\right)$ \\
\midrule
\multicolumn{4}{@{}l}{\emph{Optional local channels (add to \textsc{ATLAS} when enabled):}} \\
\quad$+\,$\textsc{NF}
& $+\,O(\|A\|_0D)$
& \multicolumn{2}{c}{input width $+D$} \\
\quad$+\,$\textsc{LPF}
& $+\,O(H\|A\|_0C)$
& \multicolumn{2}{c}{input width $+HC$} \\
\bottomrule
\end{tabular}
\vspace{-3em}
\end{table}

\begin{table}[!t]
\centering
\scriptsize
\setlength{\tabcolsep}{2.2pt}
\renewcommand{\arraystretch}{1.05}
\caption{Node classification performance (\%) $\pm$ standard deviation on
large non-homophilous graphs. $h_e$ denotes the edge homophily
ratio. Cells in \cellcolor{yellow!20}{yellow} indicate the
best score per dataset (and best Avg. Rank); \cellcolor{secondbest}{blue}, the second-best. ATLAS-NF-LPF-DIR is the full directional
model; on undirected graphs, it
reduces to ATLAS-NF-LPF.}
\label{tab:large_scale_heterophily}
\vspace{-1mm}

\resizebox{\columnwidth}{!}{%
\begin{tabular}{l c c c c c c c}
\toprule
Model
& \shortstack{Penn94 \\ \scriptsize $0.470$}
& \shortstack{Pokec \\ \scriptsize $0.445$}
& \shortstack{Arxiv-Y. \\ \scriptsize $0.222$}
& \shortstack{Snap-P. \\ \scriptsize $0.073$}
& \shortstack{Genius \\ \scriptsize $0.618$}
& \shortstack{Twitch-G. \\ \scriptsize $0.545$}
& \shortstack{Rank \\ \scriptsize $\downarrow$} \\
\midrule
MLP
& 73.61$\pm$0.40 & 62.37$\pm$0.02 & 36.70$\pm$0.21
& 31.34$\pm$0.05 & 86.68$\pm$0.09 & 60.92$\pm$0.07 & 6.50 \\
\midrule
GCN
& 82.47$\pm$0.27 & 75.45$\pm$0.17 & 46.02$\pm$0.26
& 45.65$\pm$0.04 & 87.42$\pm$0.37 & 62.18$\pm$0.26 & 4.00 \\
GAT
& 81.53$\pm$0.55 & 71.77$\pm$6.18 & 46.05$\pm$0.51
& 45.37$\pm$0.44 & 55.80$\pm$0.87 & 59.89$\pm$4.12 & 5.50 \\
\midrule
LINKX
& \cellcolor{yellow!20}\textbf{84.71$\pm$0.52}
& 82.04$\pm$0.07
& \cellcolor{secondbest}56.00$\pm$1.34
& \cellcolor{secondbest}61.95$\pm$0.12
& 90.77$\pm$0.27
& \cellcolor{yellow!20}\textbf{66.06$\pm$0.19}
& \cellcolor{secondbest}1.67 \\
ACM-GCN
& 82.52$\pm$0.96 & 63.81$\pm$5.20 & 47.37$\pm$0.59
& 55.14$\pm$0.16 & 80.33$\pm$3.91 & 62.01$\pm$0.73 & 4.17 \\
GPR-GNN
& 81.38$\pm$0.16 & 78.83$\pm$0.05 & 45.07$\pm$0.21
& 40.19$\pm$0.03 & 90.05$\pm$0.31 & 61.89$\pm$0.29 & 4.83 \\
\midrule
\textbf{ATLAS}
& 75.02$\pm$0.31 & 64.53$\pm$0.14 & 37.27$\pm$0.15
& 32.88$\pm$0.08 & 86.69$\pm$0.08 & 59.51$\pm$0.36 & \textemdash \\
\textbf{ATLAS-NF}
& 76.38$\pm$0.55 & 68.09$\pm$0.02 & 44.48$\pm$0.17
& 46.23$\pm$0.08 & 89.16$\pm$0.06 & 59.94$\pm$0.29 & \textemdash \\
\textbf{ATLAS-LPF}
& 81.96$\pm$0.70 & 82.04$\pm$0.05 & 49.20$\pm$0.31
& 51.36$\pm$0.08 & 90.34$\pm$0.18 & 65.30$\pm$0.18 & \textemdash \\
\textbf{ATLAS-NF-LPF}
& 82.60$\pm$0.55 & 82.22$\pm$0.03 & 51.22$\pm$0.35
& 57.90$\pm$0.18 & 91.58$\pm$0.10 & 65.43$\pm$0.16 & \textemdash \\
\textbf{ATLAS-NF-LPF-DIR}
& \cellcolor{secondbest}82.60$\pm$0.55
& \cellcolor{yellow!20}\textbf{82.22$\pm$0.03}
& \cellcolor{yellow!20}\textbf{60.03$\pm$0.33}
& \cellcolor{yellow!20}\textbf{66.67$\pm$0.15}
& \cellcolor{yellow!20}\textbf{91.58$\pm$0.10}
& \cellcolor{secondbest}65.43$\pm$0.16
& \cellcolor{yellow!20}\textbf{1.33} \\
\bottomrule
\end{tabular}%
}
\vspace{-1mm}
\end{table}

\paragraph {\bf Q2: What Local Channels Recover.}
\label{sec:q2}
 We test whether local channels such as NF, LPF, and the directional extension can provide accuracy when global community detection fails.

{\em Roman-Empire.}
ATLAS and MLP score identically (66.22, Table~\ref{tab:results_transposed_full}). The partition adds nothing over attributes on this role-labeled graph. However, the local channels recover most of that gap; NF 78.10 (+11.88), LPF 70.22 (+4.00), and 79.04 together. This is a 12.8-point gain the community embedding cannot express by construction, since every co-block node shares one value. 

{\em Large heterophilous graphs.}
The pattern persists at million-node scale (Table~\ref{tab:large_scale_heterophily}). ATLAS-NF-LPF trails LINKX on the two directed graphs, Arxiv-Year and Snap-Patents, by 4.78 and 4.05 points. Citations point from newer work to older, so the label rises monotonically along edge direction and the class signal lives in orientation. Restoring direction, via the gated operator $\mathcal{T}_\alpha$, turns the deficit into a lead: Arxiv-Year 60.03 (+4.03) and Snap-Patents 66.67 (+4.72), the best average rank in the suite (1.33 vs.\ 1.67). 

\paragraph{\bf Q3: Efficiency and Scalability on Large Graphs.} Table~\ref{tab:exp-sota} and Table~\ref{tab:large_scale_heterophily} show ATLAS remains competitive at million-node scale. 

On negative-bias graphs, where topology misleads, ATLAS's adaptive search discards the community channel and recovers MLP accuracy exactly (65.46\% on Yelp, 82.04\% on AmazonProducts). This is consistent with Theorem~\ref{thm:2}, since $\Delta\epsilon_n$ exceeds $I(Y;S\mid B)$ on these graphs. On label-aligned graphs, the community channel alone gives a strong baseline without message passing (95.74\% on Reddit, 79.98\% on ogbn-products), and ATLAS-LPF or ATLAS-NF-LPF add further gains when local signal is complementary, reaching a best-in-class 52.80\% on Flickr. Together with its tied-best 4.40 average rank and lower variance across dataset types, this makes ATLAS's channel-selection behavior the source of its robustness when structural bias is unknown ahead of time.

{\em Convergence}
Training loss decreases smoothly and validation performance plateaus early with a small train-validation gap and no late-epoch degradation (Figure~\ref{fig:convergence-landscape}).

{\em Efficiency.}
Table~\ref{tab:complexity} confirms the $T$-resolution community search is a one-time preprocessing cost of $O(T\|A\|_0)$; training is otherwise MLP-like and inference is adjacency-free on features of dimension $D + Td_c$ (timings in Appendix Table~\ref{tab:runtime}).

\section{Structural Bias Effects on  Refinement}
\label{refinement}

We quantify community refinement through the minimum modularity threshold $Q_{min}$. Large $Q_{min}$
retains only coarse communities, while lowering $Q_{min}$ progressively incorporates medium and fine-grained communities for multi-scale representation. We define {\em structural bias} as how effectively a graph's community structure supports classification. We identify three distinct classes.

\begin{itemize}[leftmargin=1.2em]
  \item \textbf{High structural bias} (e.g., Cora, Tolokers): Communities strongly align with labels. Coarse communities at large $Q_{\min}$
already carry substantial signal, and incorporating finer communities steadily improves performance until saturation (Figure~\ref{fig:gap_combined}, top/bottom left). Both GCN and ATLAS significantly outperform MLP.

\item \textbf{Low structural bias} (e.g., Amazon-Ratings, Chameleon-Filtered, Flickr, Squirrel-Filtered): Weak label-community alignment limits structural signal. Coarse communities capture most available information, and adding finer resolutions yields only modest gains. ATLAS moderately improves over MLP while GCN shows minimal or no improvement (Figure~\ref{fig:gap_combined}~\footnote{"Best" annotations report the accuracy observed across
a single-run sweep rather than 
averaged over multiple sweeps. }, top/bottom center).
\item \textbf{Negative structural bias} (e.g., Actor, Roman-Empire): 
Community structure actively misleads classification. Finer communities inject noise, causing performance to deteriorate as $Q_{min}$ decreases (Figure~\ref{fig:gap_combined}, top/bottom right).
\end{itemize}

\begin{figure*}[t]
    \centering
    \setlength{\tabcolsep}{6pt}
    \renewcommand{\arraystretch}{1.0}

    \begin{tabular}{ccc}

    \begin{minipage}{0.30\textwidth}
        \centering
        \includegraphics[width=\textwidth]{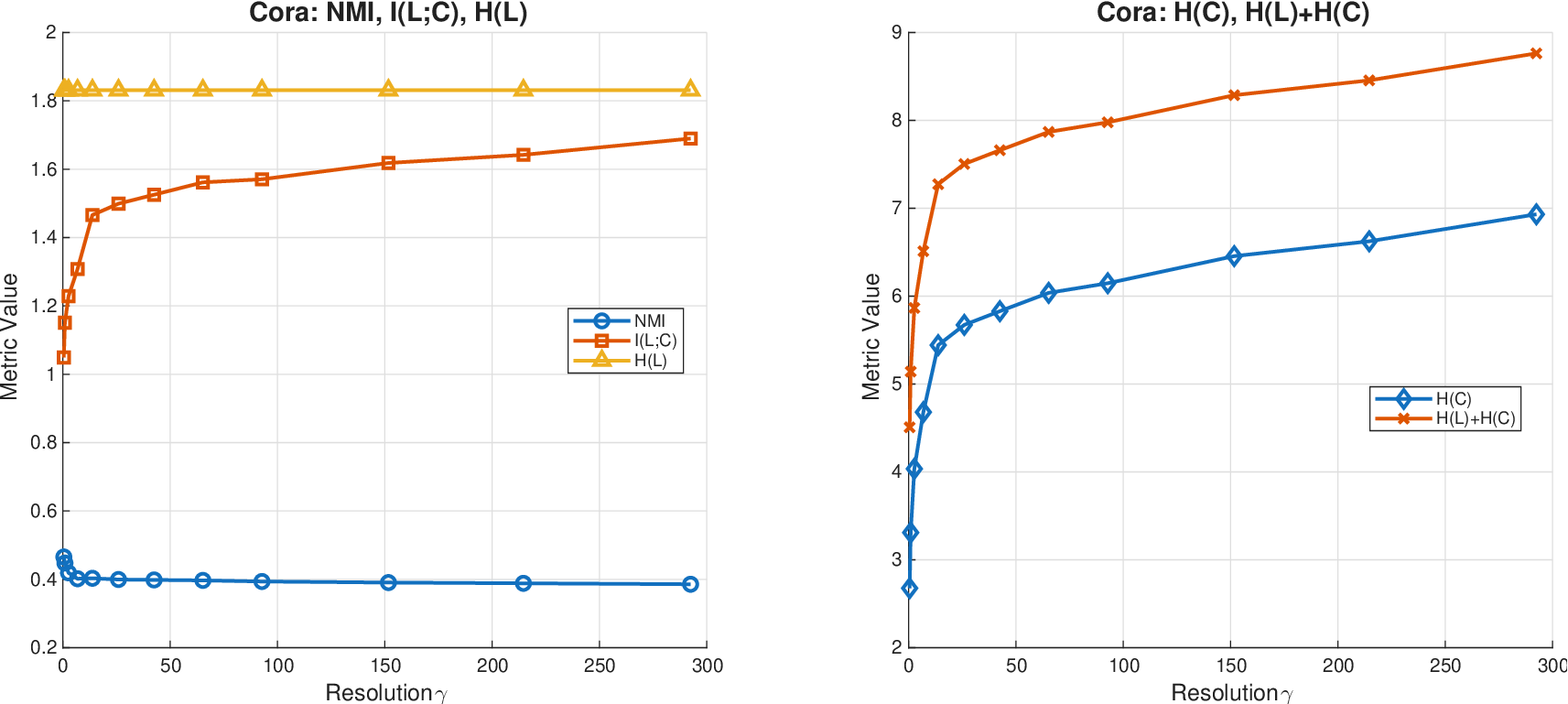}
        \caption*{(a) Cora}
        \vspace{0.15cm}
        \includegraphics[width=\textwidth]{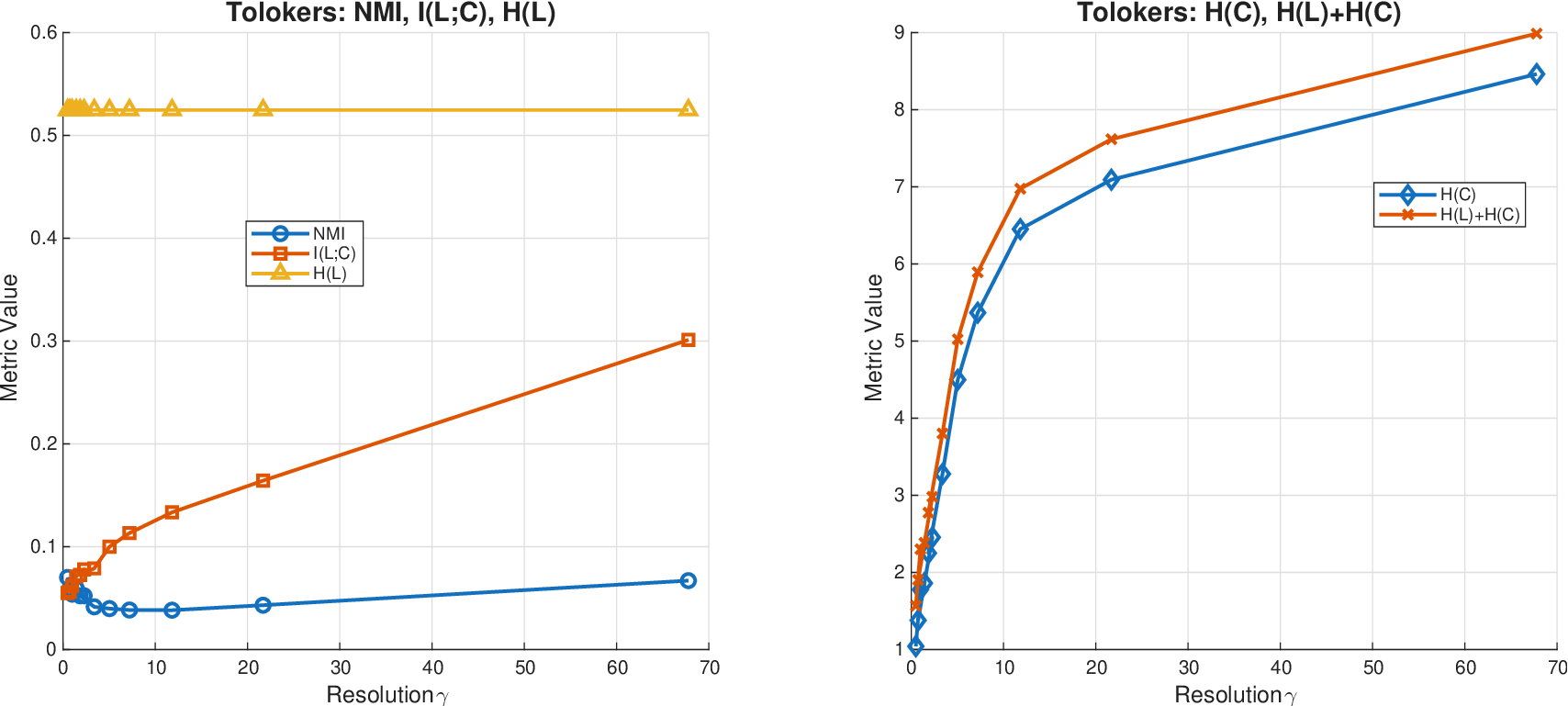}
        \caption*{(b) Tolokers}
    \end{minipage}
    &
    \begin{minipage}{0.30\textwidth}
        \centering
        \includegraphics[width=\textwidth]{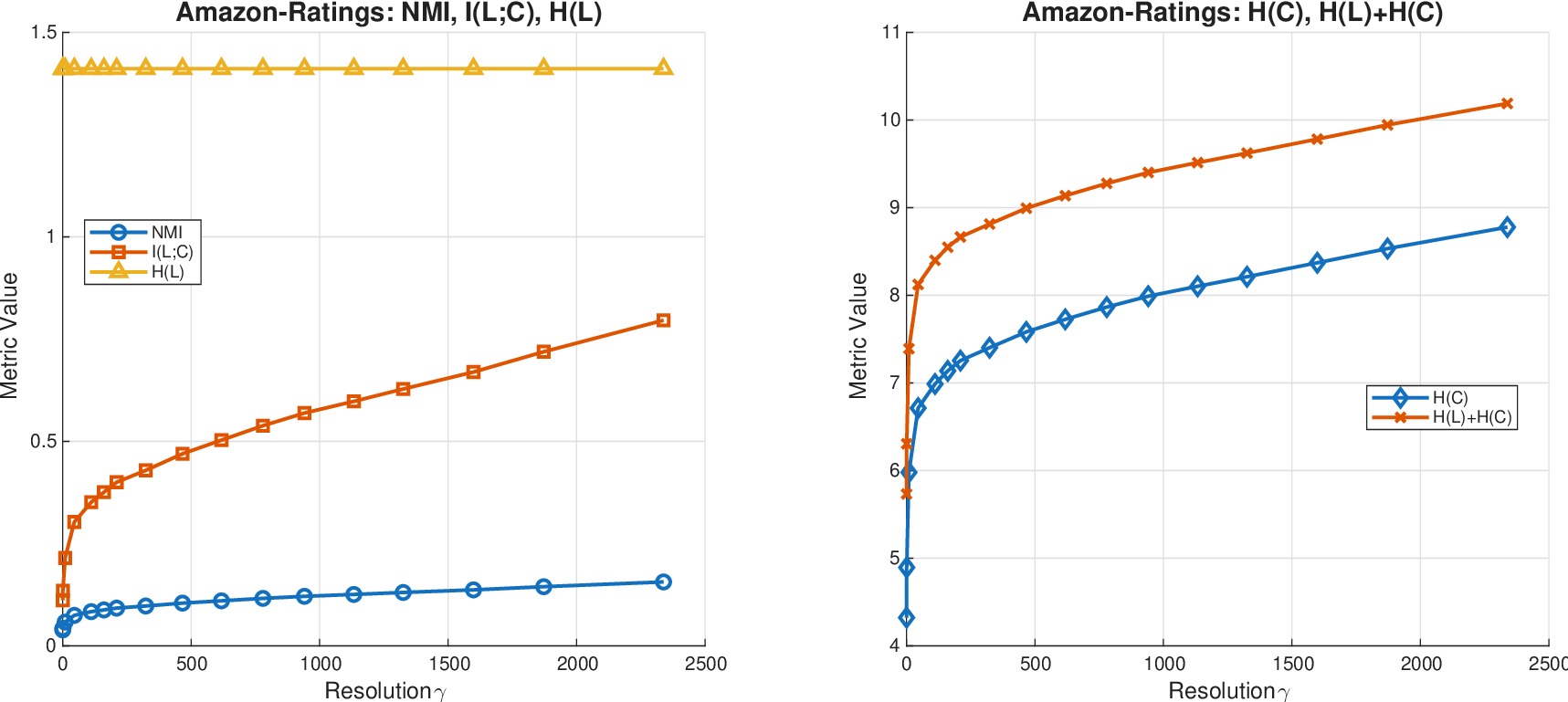}
        \caption*{(c) Amazon-Ratings}
        \vspace{0.15cm}
        \includegraphics[width=\textwidth]{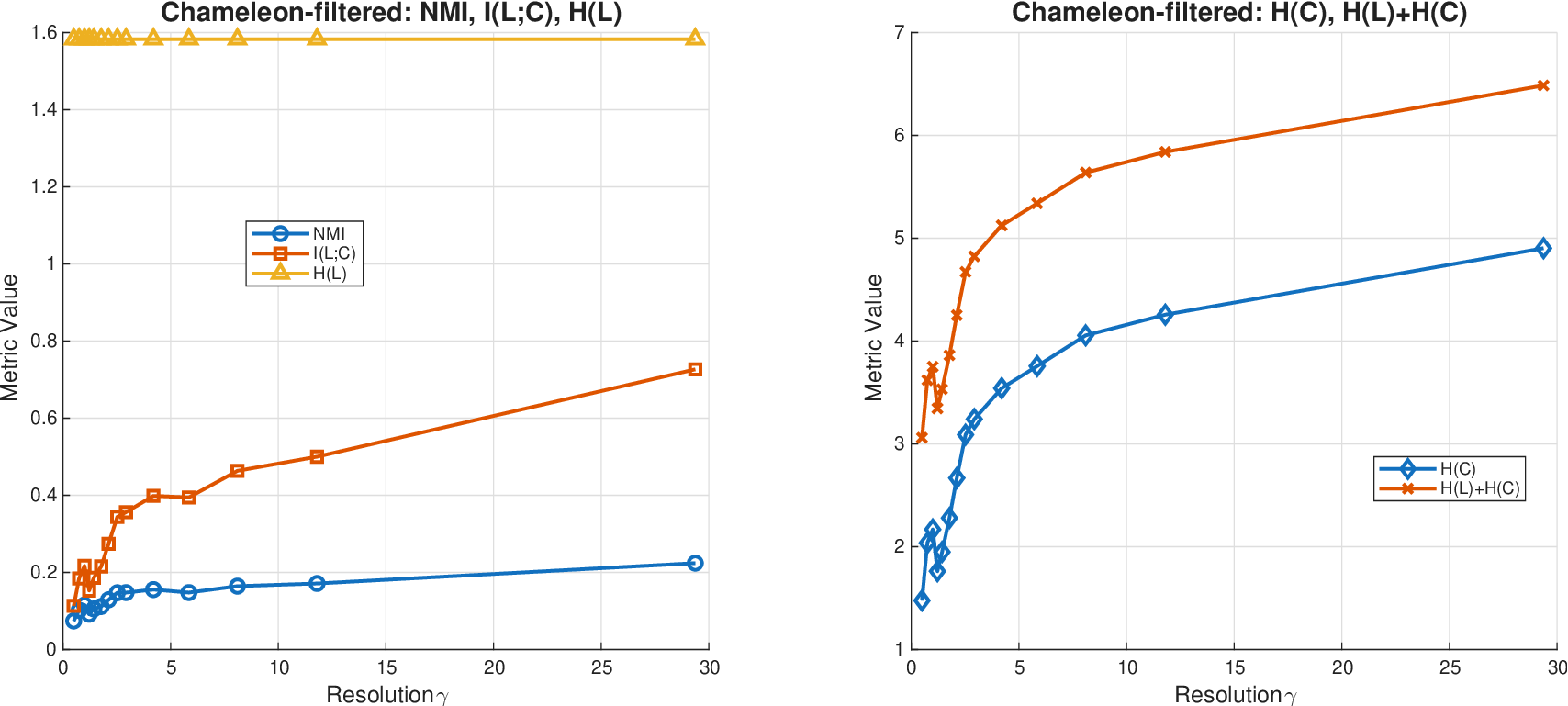}
        \caption*{(d) Chameleon-Filtered}
    \end{minipage}
    &
    \begin{minipage}{0.30\textwidth}
        \centering
        \includegraphics[width=\textwidth]{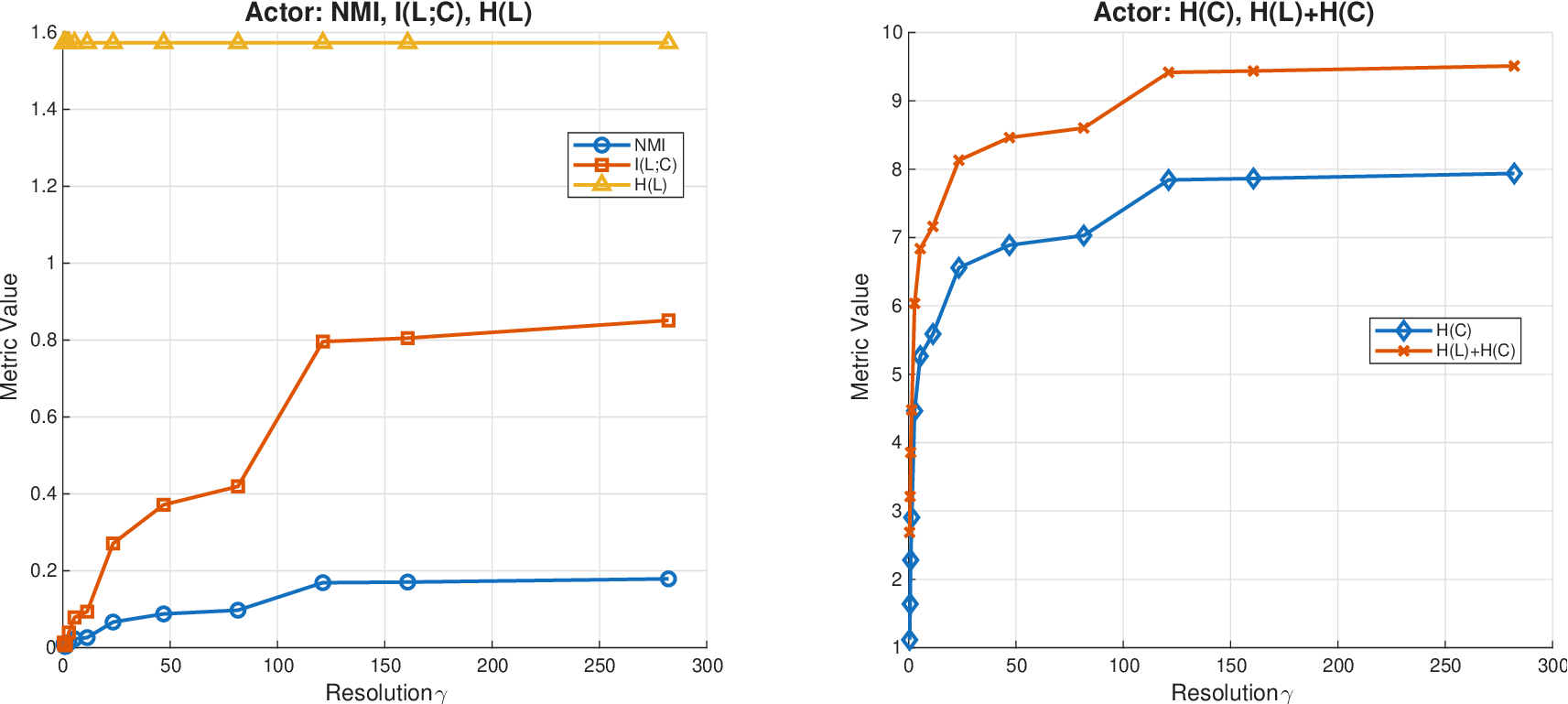}
        \caption*{(e) Actor}
        \vspace{0.15cm}
        \includegraphics[width=\textwidth]{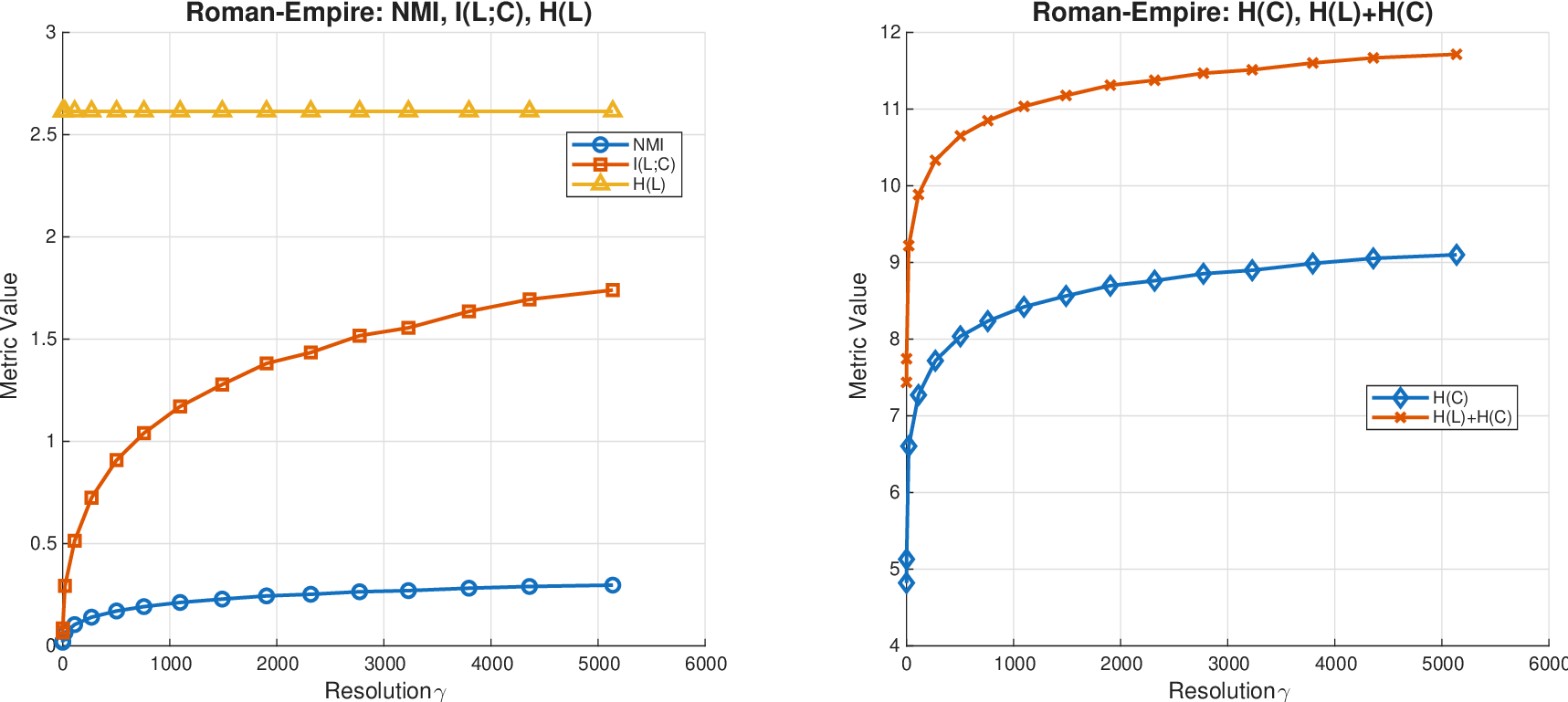}
        \caption*{(f) Roman-Empire}
    \end{minipage}

    \end{tabular}

    \caption{
    NMI, mutual information, and entropy dynamics across resolutions.
       High structural bias (Cora and Tolokers, left);
      low structural bias  (Amazon and Chameleon-Filtered, middle);
     negative structural bias  (Actor and Roman-Empire, right).
    }
    
    \label{fig:merged_structural_bias_panels}

\end{figure*}

{\bf Experimental Validation of Theory.} Figure~\ref{fig:merged_structural_bias_panels} demonstrates how community refinement behavior aligns with our NMI theory (Section~\ref{sec:theory}).  As resolution parameter $\gamma$
 increases, communities become progressively refined, and consistent with Lemmas ~\ref{lem:2}-~\ref{lem:3}, both mutual information 
$I(L;C)$ and entropy $H(C)$
grow monotonically across all datasets. Theorem~\ref{thm:3}
shows why NMI exhibits distinct patterns across structural-bias regimes. On high-bias graphs, NMI forms an interior peak where optimal label-community alignment occurs; on low-bias graphs, NMI remains low and relatively flat because gains in $I(L;C)$
fail to outpace entropy growth, indicating weak structural signal; on negative-bias graphs, NMI rises slowly from near-zero and plateaus at modest levels, reflecting fundamental misalignment between community structure and labels.

\textbf{Illustrative Example.}
Table~\ref{tab:minQ_pairs_with_acc} in  Appendix illustrates this mechanism on Cora. Accuracy rises from 76.61\% (feature-only MLP) to a peak of 88.40\% as $Q_{\min}$ is lowered. Additional coarse-to-fine community resolutions are admitted, then degrades slightly once fragmented, high-resolution communities are included as noise.

\begin{figure}[!t]
\centering
\includegraphics[width=1\columnwidth,height=.25\textheight,keepaspectratio]{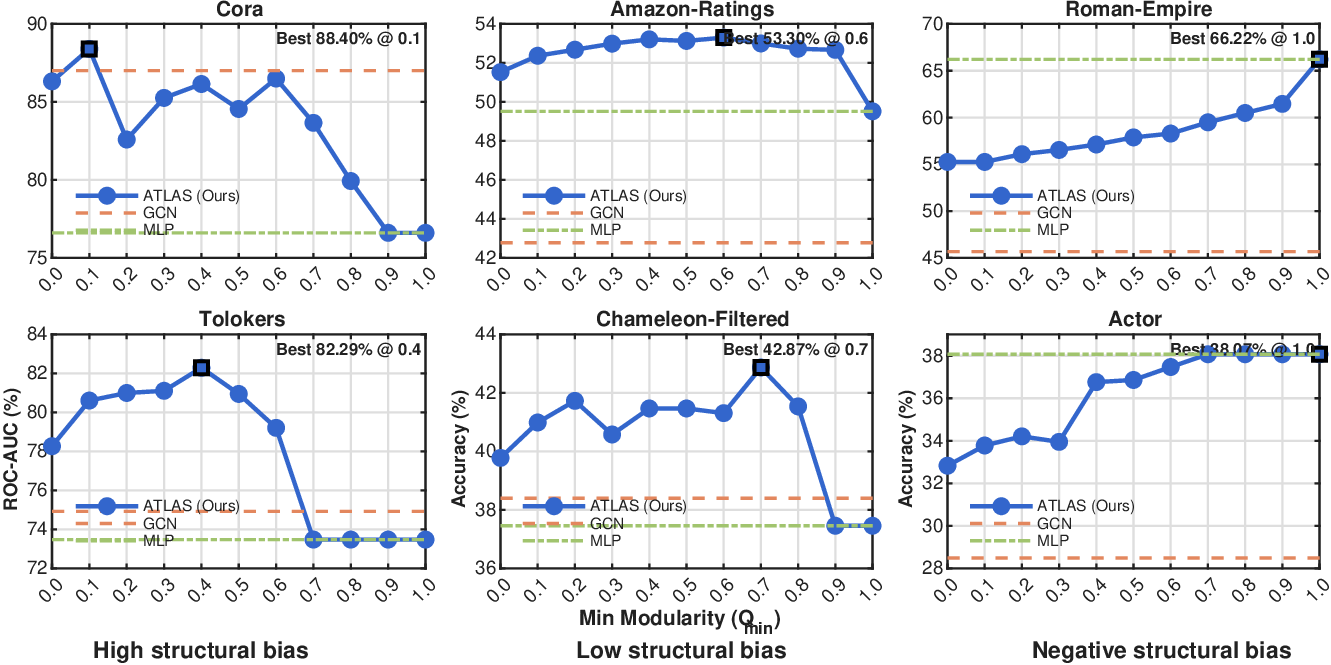}

\caption{
Effect of cumulatively adding community-derived features as the minimum modularity threshold
$Q_{\min}$ is lowered, for high structural bias graphs (left), low structural bias graphs
(middle), and negative structural bias graphs (right)
}

\label{fig:gap_combined}


\end{figure}

\begin{figure}[t]
  \centering

  \includegraphics[width=1\columnwidth,height=.15\textheight,keepaspectratio]{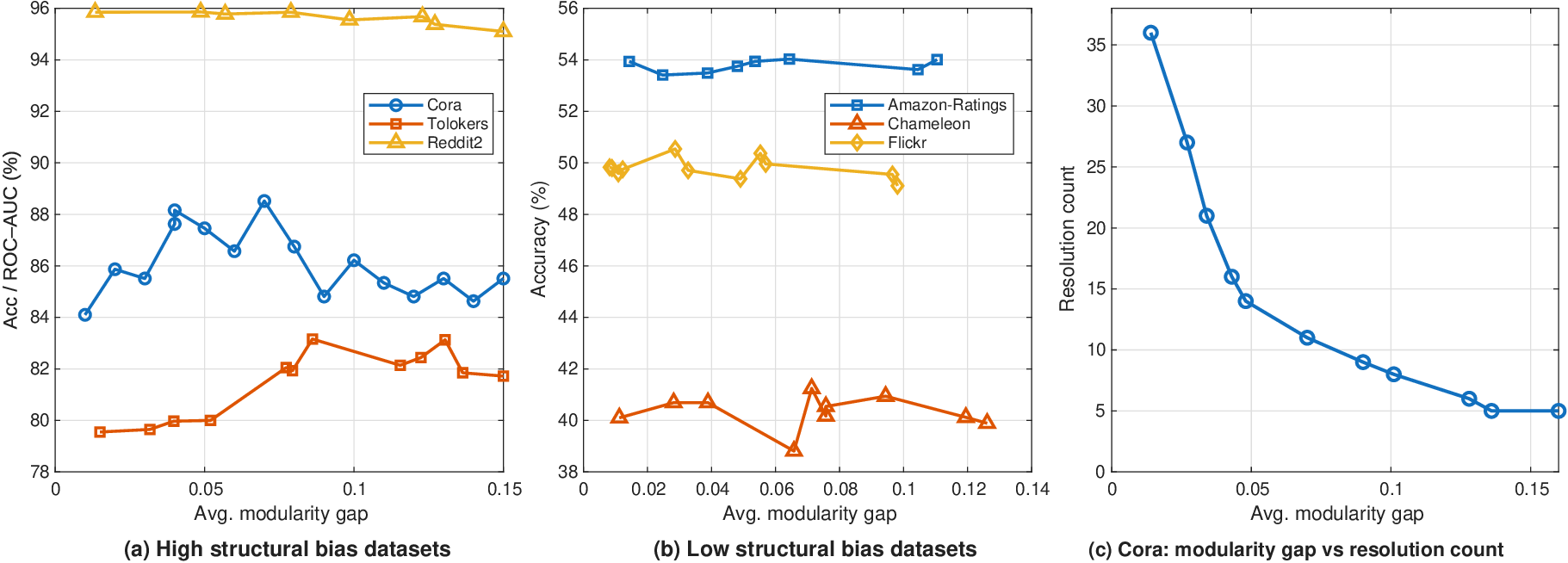} 
  \caption{
    \textbf{(a–b)} Accuracy/ROC–AUC vs.\ average modularity gap on high- and low-structural-bias datasets.
    \textbf{(c)} Resolution count vs.\ modularity gap for Cora.
  }
  \label{fig:three_panel_combined}
  \vspace{-1em}
\end{figure}

\section{Ablation Study}

\textbf{Effect of Modularity Gap and Resolution on Performance.}
The modularity gap shapes refinement behavior by determining how many Louvain resolutions are selected and their predictive value. Small average gaps retain many closely-spaced resolutions, while large gaps leave only a few widely-separated ones, trading off redundant partitions against overly-coarse structure.

On high structural-bias graphs (Figure~5(a)), accuracy follows an inverted-U in the modularity gap for Cora and Tolokers, peaking around $gap \approx 0.05$--$0.08$ before declining at larger gaps, while Reddit remains consistently high (94--96\%) with only a slight drop at large gaps. This pattern is explained in Figure~5(c), which plots resolution count against modularity gap for Cora; very small gaps retain $\sim$36 nearly-redundant resolutions that inject noise and cap accuracy near 84\%, a moderate gap ($\approx 0.05$--$0.07$) retains a compact set of $\sim$14-16 resolutions capturing distinct granularities and yielding the peak $\sim$88\% accuracy, large gaps ($>0.15$) leave fewer than 5 coarse resolutions, insufficient to exploit the structural signal.

On low structural-bias graphs (Figure~5(b)), the pattern is markedly flatter: Amazon-Ratings stays within 53--54\% and Flickr within 49--51\% across all gap values, with Chameleon showing the most fluctuation (39--42\%). Since community structure carries minimal label information on these graphs, varying the gap mainly changes how many resolutions are retained without making them substantially more predictive, so community features yield only modest gains over feature-only baselines regardless of gap setting.

The resolution-count mechanism in Figure~5(c) explains the inverted-U accuracy pattern for high-bias graphs. Optimal performance requires balancing sufficient structural information (enough resolutions) against feature redundancy (not too many near-duplicate resolutions), while on low-bias graphs this balance has little effect on accuracy since the underlying structural signal is weak.

\vspace{-1 em}
\section{Related Work}

\textbf{Scalable and Heterophily-Aware GNNs.} While sampling techniques \cite{chen2018fastgcn,hamilton2017inductive,zou2019layer,chiang2019cluster} and decoupled architectures (e.g., SGC, SIGN, GAMLP) \cite{rossi2020sign,wu2019simplifying,zhang2022gamlp} address memory bottlenecks via stochasticity or precomputed feature diffusion, they frequently require separate architectural mechanisms to handle heterophily via neighbor reweighting, self-feature preservation, or higher-order propagation \cite{abu-el-haija2019mixhop,bo2021beyond,chien2022adaptive,he2021bernnet,li2022finding,zhang2022graph}. ATLAS instead avoids repeated, per-epoch message passing, using a
small, fixed number of one-time propagation steps during
preprocessing.


\textbf{Community-Aware Graph Learning.} Prior frameworks capture this by jointly inferring communities and embeddings \cite{sun2019vgraph}, constructing community-derived features \cite{Kaminski2024CommunityFeatures}, or performing community-guided edge rewiring \cite{rubio-madrigal2025gnns}. ATLAS instead treats multi-resolution community assignments strictly as MLP input features, exploiting topological structure without propagation steps.

\textbf{Compatibility-Aware Propagation.} A parallel line of work leverages explicit class-compatibility matrices to inform message passing or label propagation, e.g., H2GCN \cite{zhu2020beyond}, CPGNN \cite{zhu2021cpgnn}, CLP \cite{zhong2022clp}, and CMGNN \cite{zheng2025cmgnn}. ATLAS instead utilizes degree-normalized label priors as precomputed features, shifting compatibility handling as a  nearly propagation-free inference pipeline.

\section{Conclusion and Limitations}
\label{sec:conclusion}

ATLAS is a nearly propagation-free graph learning framework that injects topology via multi-resolution community, one-hop neighbor features, and  label priors, training a compact MLP over the concatenated representation. ATLAS achieves competitive or superior accuracy relative to homophily-oriented GNNs, heterophily-oriented models, and scalable baselines, with stable convergence and adjacency-free inference once preprocessing completes. Structural-bias analysis demonstrates refinement helps when community structure aligns with labels and hurts when it does not, giving an interpretable view of when topology benefits classification.

Two limitations temper these results. First, our theory assumes idealized nested partitions which  real community detection may not satisfy. Thus alignment is measured via NMI rather than task-level accuracy guarantees. Second, the multi-resolution search runs $T$ separate Louvain optimizations and being dependent on Louvain's sensitivity to initialization and seed selection. Future work includes extending the theory to approximate refinements and connecting alignment to learning dynamics, using hierarchical methods to compute multiple granularities in fewer passes. 
\newpage
\textbf{Ethics Statement.} We confirm that we have adhered to the KDD Code of Ethics.

\textbf{Use of Generative AI.} We have used generative AI to polish the writing, and to check that the proofs of the theorem and lemma are correct and concise.

\textbf{Reproducibility Statement.} Our source code is available at \url{https://github.com/atlaspaper16/ATLAS}.

\bibliographystyle{ACM-Reference-Format}
\bibliography{citation_cleaned}
\newpage
\appendix

\section{Appendix}

\paragraph{Compute environment.}
All experiments were run on a server with 1$\times$ NVIDIA A40 (45\,GiB) GPU, 32 vCPUs, 2$\times$ Intel Xeon Silver 4309Y @ 2.80\,GHz, and 503\,GiB RAM.\\
Software stack: Python~3.10.18; PyTorch~2.4.0{+}cu124 (CUDA~12.4); PyTorch~Geometric~2.6.1.

\subsection{Definitions and Proofs}
\label{app:community_defs}

This subsection defines the modularity-based community detection terms that
underpin our multi-resolution community extraction.

\paragraph{Modularity.}
Given a partition $C=\{C_1,\ldots,C_K\}$ with node assignments $c_i\in\{1,\ldots,K\}$,
\emph{modularity} measures how much denser the intra-community connections are
than expected under a degree-preserving null model:
\begin{equation}
    Q \;=\; \frac{1}{2m}\sum_{i,j}\Big(A_{ij}-\frac{k_i k_j}{2m}\Big)\,\delta(c_i,c_j),
\end{equation}
where $A_{ij}$ is the adjacency matrix, $k_i$ is the degree of node $i$,
$m=|E|$ is the number of edges, and $\delta(c_i,c_j)=1$ if $c_i=c_j$ (else $0$).
Higher $Q$ indicates stronger community structure.

\paragraph{Resolution parameter.}
Louvain introduces a \emph{resolution} $\gamma>0$ to control granularity by
reweighting the null-model term:
\begin{equation}
    Q(\gamma)\;=\; \frac{1}{2m}\sum_{i,j}\Big(A_{ij}-\gamma\frac{k_i k_j}{2m}\Big)\,\delta(c_i,c_j).
\end{equation}
Smaller $\gamma$ favors coarser partitions, while larger $\gamma$ typically yields
finer (more, smaller) communities, providing partitions at multiple granularities across $\gamma$.
We denote the resulting partition and modularity by $C(\gamma)$ and $Q^{(\gamma)}$,
with assignment vector $\mathbf{c}^{(\gamma)}$.

\paragraph{Modularity gap.}
For two values $\gamma_1<\gamma_2$ that are adjacent after sorting the set of $\gamma$ values, the \emph{modularity gap}
quantifies the change in community quality:
\begin{equation}
    \Delta Q(\gamma_1,\gamma_2)\;=\;\big|Q^{(\gamma_2)}-Q^{(\gamma_1)}\big|.
\end{equation}
Large gaps indicate rapid structural changes between scales and motivate inserting
intermediate resolutions; small gaps suggest the refinement has stabilized.

All entropies and mutual informations below are computed under the
empirical (uniform) distribution over nodes. Two facts are used
repeatedly, so we prove them first.

\paragraph{Fact 1: Conditional MI is non-negative, i.e. for any $Y,S,B$: $I(Y;S\mid B) \ge 0$.}
\label{fact:nonneg}
\begin{proof}
Fix $B=b$. Then
\[
I(Y;S\mid B{=}b)
= \sum_{y,s} p(y,s\mid b)
  \log\frac{p(y,s\mid b)}{p(y\mid b)p(s\mid b)}
\]
is a KL divergence, which is non-negative by Gibbs' inequality.
Averaging over $b$ preserves non-negativity.
\end{proof}

\paragraph{Fact 2: Chain rule. For any $Y,B,S$: $I(Y;B,S) = I(Y;B) + I(Y;S\mid B)$.}
\label{fact:chain}

\begin{proof}
Both sides equal $H(Y) - H(Y\mid B,S)$ by expanding the definitions of
mutual and conditional mutual information.
\end{proof}

\subsection{Lemma 1: channels are complementary}
\emph{Idea:} adding a channel never destroys information, and the gain
from a second channel is exactly the leftover information it carries
once the first is known.

\begin{proof}
Fact~2 gives $I(Y;B,S)-I(Y;B)=I(Y;S\mid B)$, which is
the stated equality. Fact~1 makes this difference
$\ge 0$, so $I(Y;B,S)\ge I(Y;B)$. The same argument with $B,S$
swapped gives $I(Y;B,S)\ge I(Y;S)$. Together these give
$I(Y;B,S)\ge\max\{I(Y;B),I(Y;S)\}$.
\end{proof}

\subsection{Theorem 1: adding a channel never hurts the ideal classifier}
\emph{Idea:} the Bayes log-loss risk of a representation $U$ is just
$H(Y\mid U)$, so risk and mutual information move in lockstep.

\begin{proof}
Write $R^*(U):=H(Y\mid U)$ for the Bayes log-loss risk under $U$; this
is the risk of predictor $P(Y\mid U)$, which minimizes expected
log-loss by Gibbs' inequality. Then
\[
R^*(B) - R^*(B,S) = I(Y;S\mid B) \ge 0,
\]
using Fact~1. So $R^*(B,S)\le R^*(B)$ always, and
strictly less whenever $I(Y;S\mid B)>0$.
\end{proof}

\subsection{Theorem 2: finite-sample inclusion criterion}
\emph{Idea:} with finite data, a channel also costs something to
learn. Split expected risk into a Bayes part and an estimation part,
and the two channel comparisons combine cleanly.

\begin{proof}
For any $U$, define the excess (estimation) risk
\[
\epsilon_n(U) := \mathbb{E}\big[R_U(\hat f_U)\big] - R^*(U) \;\ge\; 0.
\]
Apply this at $U=B$ and $U=(B,S)$ and subtract:
\[
\mathbb{E}_S\big[R_{B,S}(\hat f_{B,S}) - R_B(\hat f_B)\big]
= \underbrace{\big(R^*(B,S)-R^*(B)\big)}_{=\,-I(Y;S\mid B)}
+ \underbrace{\big(\epsilon_n(B,S)-\epsilon_n(B)\big)}_{=:\,\Delta\epsilon_n(S\mid B)}.
\]
The left side is negative---i.e.\ adding $S$ helps---exactly when
\[
I(Y;S\mid B) > \Delta\epsilon_n(S\mid B).
\]
As $n\to\infty$, $\epsilon_n\to 0$ for any fixed hypothesis class, so
$\Delta\epsilon_n\to 0$ and the criterion reduces to Theorem~1.
\end{proof}

\subsection{Lemma 2: finer partitions know at least as much as coarse partitions}
\emph{Idea:} a coarse partition is recoverable from a fine one, so it
adds nothing once the fine one is known.

\begin{proof}
If $C'\preceq C$ then $C=g(C')$ for a fixed map $g$, so
$H(C\mid C')=0$. The chain rule (twice, conditioned on $Z$) gives
\[
I(Y;C'\mid Z) = I(Y;C\mid Z) + I(Y;C'\mid C,Z).
\]
The last term is non-negative by Fact~1, so
\[
I(Y;C'\mid Z) \ge I(Y;C\mid Z).
\]
\end{proof}

\subsection{Lemma 3: finer partitions cost more entropy}
\emph{Idea:} more, smaller communities need more bits to describe, up
to the trivial ceiling of one bucket per node.

\begin{proof}
As above, $H(C\mid C')=0$, so by the entropy chain rule
\[
H(C') = H(C) + H(C'\mid C) \;\ge\; H(C).
\]
If some block of $C$ is split into $\ge 2$ nonempty pieces of $C'$,
then $H(C'\mid C{=}c)>0$ for that block, so $H(C'\mid C)>0$ and the
inequality is strict. Since any partition of $n$ nodes has at most
$n$ blocks, $H(C)\le\log n$ always, with equality only for the
all-singletons partition.
\end{proof}

\subsection{Theorem 3: when refining improves NMI}
\emph{Idea:} refining trades an information gain $\Delta I \ge 0$
against an entropy cost $\Delta H > 0$; NMI improves only if that
trade beats the current alignment level.

\begin{proof}
Let $I,I',H,H'$ denote $I(C;Y),I(C';Y),H(C),H(C')$, and
$H_Y := H(Y)$. Then
\[
\mathrm{NMI}(C',Y) > \mathrm{NMI}(C,Y)
\iff
\frac{I'}{H'+H_Y} > \frac{I}{H+H_Y}.
\]
Cross-multiply and substitute $I'=I+\Delta I$, $H'=H+\Delta H$:
\[
I'(H+H_Y) > I(H'+H_Y)
\iff
\Delta I\,(H+H_Y) > I\,\Delta H.
\]
Divide by $\Delta H>0$:
\[
\frac{\Delta I}{\Delta H} > \frac{I}{H+H_Y} = \frac{\mathrm{NMI}(C,Y)}{2}.
\]
\end{proof}


\subsection{Lemma 4: variance of the one-hop average}

\noindent\emph{Idea:} averaging a node's neighbors mixes real signal
(same class, shared across the neighborhood) with correlated mismatch
noise; the two pieces are independent, so their variances add.

\smallskip
\noindent We model match status at the \emph{neighborhood} level rather
than per edge, since averaging over $d_i$ independent match indicators
would instead give the (weaker, exact) bound
$(\sigma^2 + (1-\rho)v)/d_i$; the shared-indicator model is the more
conservative choice under correlated heterophily and is the one used
throughout Theorem~4 and its proof.

\begin{proof}
Write $X_{\mathrm{NF},i} - \mu_{Y_i} = M\bar{\eta} + (1-M)\Delta$, where
$\bar{\eta}$ is the mean of $d_i$ i.i.d.\ noise terms $\eta_j$, $M \sim
\mathrm{Bernoulli}(\rho)$ marks a same-class neighborhood, and $\Delta$ is
the shared mismatch offset with mean $0$ and variance $v$. These pieces
are mutually independent, so
\[
\mathrm{Var}(M\bar\eta) = \rho^2\,\frac{\sigma^2}{d_i}, \qquad
\mathrm{Var}\big((1-M)\Delta\big) = (1-\rho)\,v,
\]
using $\mathrm{Var}(\bar\eta) = \sigma^2/d_i$ and $M^2 = M$. Adding the two
independent, mean-zero pieces gives
\[
\mathrm{Var}(X_{\mathrm{NF},i} \mid Y_i) = \rho^2\,\frac{\sigma^2}{d_i}
+ (1-\rho)\,v. \qedhere
\]
\end{proof}

\subsection{Theorem 4: when neighbor features help}
\emph{Idea:} neighbor averaging helps once it lowers the within-class
variance below the raw attribute's variance $\sigma^2$; below a
threshold $\rho^*$ it does not.

\begin{proof}
Let $g(\rho)$ be the variance from Lemma~4 (restated in
Appendix~A.8), and $h(\rho) := \sigma^2 - g(\rho)$. Under a
Gaussian-channel approximation, $I(Y; X_{\mathrm{NF}} \mid X) > 0
\iff h(\rho) > 0$.

Since $h(0) = \sigma^2 - v < 0$ (when $v > \sigma^2$) and
$h(1) = \sigma^2(1 - 1/d_i) > 0$, and $h$ is increasing whenever
$v > 2\sigma^2/d_i$, the Intermediate Value Theorem gives a unique
root $\rho^* \in (0,1)$ with $h(\rho) > 0 \iff \rho > \rho^*$. This
proves the crossover claim. As $\rho \to 0$, $X_{\mathrm{NF},i}$
becomes independent of $Y_i$ given $X_i$ (no systematic bias in
$\Delta$), so $I(Y; X_{\mathrm{NF}} \mid X) \to 0$, consistent with
$\rho < \rho^*$.
\end{proof}

\subsection{Lemma 5: longer label walks know at least as much}
\emph{Idea:} same argument as Lemma~2, one hop at a time.

\begin{proof}
Let $S_H=(Y^{(1)},\dots,Y^{(H)})$. The chain rule gives
\[
I(Y;S_{H+1}\mid X) = I(Y;S_H\mid X) + I\big(Y;Y^{(H+1)}\mid S_H,X\big),
\]
and the last term is $\ge 0$ by Fact~1.
\end{proof}

\subsection{Theorem 5: label priors oversmooth geometrically}
\emph{Idea:} the random walk forgets its start exponentially fast, at
a rate set by the graph's spectral gap; only the ``forgotten'' part
carries usable signal.

\begin{proof}
Write $\bar P=\Pi+E$, where $\Pi=\mathbf{1}\pi^\top$ is the rank-one
projector onto the stationary distribution $\pi$. Since
$\Pi\bar P=\bar P\Pi=\Pi$, induction gives $\bar P^h=\Pi+E^h$. The
matrix $E$ has the same eigenvalues as $\bar P$ except that its
leading eigenvalue $1$ is replaced by $0$, so all its eigenvalues have
modulus $\le\lambda$ (the second-largest eigenvalue modulus of
$\bar P$). Hence there is a constant $C$ with
$\|E^h\|_{\mathrm{op}}\le C\lambda^h$.

Since $Y^{(h)}=\bar P^h Y^{(0)}$,
\[
Y^{(h)} - \underbrace{\Pi Y^{(0)}}_{\text{node-independent}}
= E^h Y^{(0)} =: R_h,
\qquad
\|R_h\| \le C\lambda^h\|Y^{(0)}\|.
\]
The first term is identical at every node, so it carries no usable
signal; a classifier absorbs it as a constant offset. For an
$L$-Lipschitz downstream readout, the risk reduction attributable to
$Y^{(h)}$ is bounded by $L\|R_h\|$, giving
\[
\big|\text{risk reduction from } Y^{(h)}\big| \;=\; O(L\lambda^h).
\]
\end{proof}

\subsection{Theorem 6: no channel is redundant}
\emph{Idea:} build three toy graphs, one per channel, where only that
channel carries signal; then glue them together so all three channels
are simultaneously needed.

\begin{proof}
\textbf{(a) Community.} Take $K\ge2$ disjoint isomorphic blocks with
i.i.d.\ attributes and no cross-block edges; let $Y$ be the block
index. Attributes, neighbor features, and label priors all look
identical across blocks, so
\[
I(Y;X_{\mathrm{NF}},Y_{\mathrm{LPF}}\mid X)=0.
\]
But the community partition recovers $Y$ exactly, so
$I(Y;\mathcal{C}\mid X,X_{\mathrm{NF}},Y_{\mathrm{LPF}}) = H(Y) > 0$.

\textbf{(b) Neighbor features.} Take one connected expander graph (a
single community at every resolution) with local same-class rate
$\rho>\rho^*$ from Theorem~4, and withhold labels so
$Y_{\mathrm{LPF}}\equiv 0$. Then $\mathcal{C}$ and $Y_{\mathrm{LPF}}$
are both constant, but $I(Y;X_{\mathrm{NF}}\mid X)>0$ by Theorem~4.

\textbf{(c) Label priors.} Take the same single-community graph but
with attributes carrying no signal ($X\perp Y$, so
$X_{\mathrm{NF}}\perp Y$ too), and a class-compatibility pattern with
non-identical rows. Then one hop of label diffusion is informative:
$I(Y;Y_{\mathrm{LPF}}\mid X,\mathcal{C},X_{\mathrm{NF}}) > 0$.

\textbf{Joint dominance.} Glue three equal, non-adjacent copies of
these graphs together, one realizing each case. With no edges between
copies, information is additive across the mixture:
\[
I(Y;\mathcal{C},X_{\mathrm{NF}},Y_{\mathrm{LPF}}\mid X)
= \tfrac13\sum_{k=1}^{3} I_{P_k}(Y;\text{channel}_k\mid X).
\]
Dropping any one channel drops exactly one term in this sum, which is
strictly positive by (a)--(c). So all three channels together
strictly beat any two.
\end{proof}

\subsection{Seed Stability of the Community Channel}
\label{app:seed}

Louvain is randomized, so both the community family the adaptive search retains
and the size $|\Gamma|$ of the retained resolution set are seed-dependent. The
dependence touches only the community channel: $X$ is fixed, and
$X_{\mathrm{NF}},Y_{\mathrm{LPF}}$ are deterministic functions of $\bar A$
and the training mask. We show that seed variation enters the inclusion
criterion of Theorem~\ref{thm:2} as an additive term on the
information side, so reseeding is governed by the same
information-versus-cost trade-off as every other channel.

\paragraph{Definitions.}
Because $|\Gamma|$ varies with the seed, comparing two runs resolution by
resolution would demand a correspondence that random reseeding does not
preserve. We therefore analyze a family $\mathcal{H}=\{C^{(1)},\dots,C^{(T)}\}$
through its meet in the refinement order, $C^{\ast}=\bigwedge_{t\le T}C^{(t)}$,
whose blocks are the nonempty intersections $\bigcap_{t}B_t$ with
$B_t\in C^{(t)}$. A node's membership vector
$\bigl(C^{(1)}(v),\dots,C^{(T)}(v)\bigr)$ determines, and is determined by, the
block of $C^{\ast}$ containing $v$; hence
$\sigma(\mathcal{H})=\sigma(C^{\ast})$ and
\begin{equation}
I(Y;\mathcal{H}\mid B)=I(Y;C^{\ast}\mid B)\quad\text{for every base }B.
\label{eq:cstar}
\end{equation}
Disagreement between a family and its rerun $\widetilde{\mathcal{H}}$ is
measured on the two meets by the variation of information~\cite{meila2007vi}
\begin{equation}
\nu^{\ast}=\mathrm{VI}(C^{\ast},\widetilde{C}^{\ast})
=H(C^{\ast}\mid\widetilde{C}^{\ast})+H(\widetilde{C}^{\ast}\mid C^{\ast}),
\label{eq:vi}
\end{equation}
which is invariant to relabeling and needs no matching of resolutions across
seeds.

\paragraph{Stability bound.}
\begin{theorem}[Family-level seed stability]
\label{thm:seedvi}
For any base representation $B$,
$\bigl|I(Y;\mathcal{H}\mid B)-I(Y;\widetilde{\mathcal{H}}\mid B)\bigr|\le
\nu^{\ast}$.
\end{theorem}

\begin{proof}
By \eqref{eq:cstar} the left-hand side equals
$\bigl|I(Y;C^{\ast}\mid B)-I(Y;\widetilde{C}^{\ast}\mid B)\bigr|$. Expanding
$I(Y;C^{\ast},\widetilde{C}^{\ast}\mid B)$ by the chain rule in both orders,
\[
I(Y;C^{\ast}\mid B)-I(Y;\widetilde{C}^{\ast}\mid B)
=I(Y;C^{\ast}\mid B,\widetilde{C}^{\ast})
-I(Y;\widetilde{C}^{\ast}\mid B,C^{\ast}).
\]
Both terms are nonnegative, and
$I(Y;C^{\ast}\mid B,\widetilde{C}^{\ast})\le
H(C^{\ast}\mid B,\widetilde{C}^{\ast})\le H(C^{\ast}\mid\widetilde{C}^{\ast})$,
with the symmetric bound for the second term. The absolute difference of two
nonnegative numbers is at most their maximum, hence at most the
sum equation~\eqref{eq:vi}.
\end{proof}

\begin{remark}
\label{rem:subadd}
If two runs retain the same number of resolutions and are compared in matched
order, then $\nu^{\ast}\le\sum_{t}\nu_t$ with
$\nu_t=\mathrm{VI}(C^{(t)},\widetilde{C}^{(t)})$: the meet is a function of the
family, and conditioning on the matched family only lowers entropy. We record
this for intuition only. The matching it presumes is exactly what random
reseeding need not preserve, which is why the analysis is carried out at the
level of $C^{\ast}$.
\end{remark}

\begin{corollary}[Stability-adjusted inclusion]
\label{cor:seedincl}
If a reference run satisfies
$I(Y;\widetilde{C}^{\ast}\mid X)>\Delta\epsilon_n(\mathcal{H}\mid X)+\nu^{\ast}$,
then $I(Y;\mathcal{H}\mid X)>\Delta\epsilon_n(\mathcal{H}\mid X)$: the community
channel passes the inclusion criterion of Theorem~\ref{thm:2} for the
family the current seed actually retained.
\end{corollary}

\begin{proof}
Theorem~\ref{thm:seedvi} at $B=X$, together with \eqref{eq:cstar}, gives
$I(Y;\mathcal{H}\mid X)\ge I(Y;\widetilde{C}^{\ast}\mid X)-\nu^{\ast}$;
combining with the hypothesis yields the claim.
\end{proof}

\begin{corollary}[Cap under informative local channels]
\label{cor:seedlocal}
For $B=(X,X_{\mathrm{NF}},Y_{\mathrm{LPF}})$,
\[
\bigl|I(Y;\mathcal{H}\mid B)-I(Y;\widetilde{\mathcal{H}}\mid B)\bigr|
\le\min\bigl\{\nu^{\ast},\,H(Y\mid X,X_{\mathrm{NF}},Y_{\mathrm{LPF}})\bigr\}.
\]
\end{corollary}

\begin{proof}
The first bound is Theorem~\ref{thm:seedvi} at this $B$. For the second, both
quantities lie in $[0,H(Y\mid B)]$, since
$I(Y;\mathcal{H}\mid B)=H(Y\mid B)-H(Y\mid B,\mathcal{H})$ and conditional
entropy is nonnegative; the absolute difference of two numbers in that interval
is at most $H(Y\mid B)$.
\end{proof}

\begin{corollary}[Seed stability of expected risk]
\label{cor:seedrisk}
Fix a base $B$ and a hypothesis class. Then
\[
\Bigl|\mathbb{E}\bigl[R_{B,\mathcal{H}}(\hat f_{B,\mathcal{H}})\bigr]
-\mathbb{E}\bigl[R_{B,\widetilde{\mathcal{H}}}(\hat f_{B,\widetilde{\mathcal{H}}})\bigr]\Bigr|
\le\nu^{\ast}
+\bigl|\Delta\epsilon_n(\mathcal{H}\mid B)
-\Delta\epsilon_n(\widetilde{\mathcal{H}}\mid B)\bigr|.
\]
\end{corollary}

\begin{proof}
Apply Theorem~\ref{thm:2} to each family over the same base and
subtract; the common term $\mathbb{E}[R_B(\hat f_B)]$ cancels, leaving the
difference of the two information terms and the difference of the two cost
terms. Bound the former by Theorem~\ref{thm:seedvi} and apply the triangle
inequality.
\end{proof}

The two summands separate the two ways a seed can matter: it can change what
the family knows about $Y$, and---through $|\Gamma|$, which sets the number of
one-hot dimensions the MLP must fit---it can change what the family costs to
use. Corollary~\ref{cor:seedincl} pads only the information side; the cost term
there is the one belonging to the family actually retained, which is why the
criterion needs no transfer of costs between seeds.

\newcommand{\pmv}[1]{{\tiny$\pm$#1}}   

\begin{table}[t]
\centering
\caption{Fixed-seed versus random-seeded preprocessing; mean $\pm$ std
over $10$ runs, $\Delta$ is random-seeded minus fixed. Each dataset
lists base ATLAS together with its best-performing local-channel
variant from Tables~\ref{tab:results_transposed_full}
and~\ref{tab:exp-sota}.}
\label{tab:seed}
\scriptsize
\setlength{\tabcolsep}{2.5pt}
\begin{tabular}{@{}llccc@{}}
\toprule
Dataset & Model & Fixed & Random & $\Delta$ \\
\midrule
Cora              & ATLAS        & 88.37\pmv{0.63} & 87.44\pmv{0.52} & $-0.93$ \\
                  & ATLAS-LPF-NF & 89.20\pmv{0.39} & 88.86\pmv{0.51} & $-0.34$ \\
\addlinespace[2pt]
Chameleon (Filt.) & ATLAS        & 42.87\pmv{3.55} & 41.66\pmv{4.01} & $-1.21$ \\
                  & ATLAS-LPF    & 42.69\pmv{4.07} & 41.10\pmv{3.46} & $-1.59$ \\
\addlinespace[2pt]
Tolokers          & ATLAS        & 82.19\pmv{0.73} & 81.78\pmv{0.64} & $-0.41$ \\
                  & ATLAS-NF     & 82.88\pmv{0.78} & 82.56\pmv{0.62} & $-0.32$ \\
\addlinespace[2pt]
Amazon-Ratings    & ATLAS        & 53.17\pmv{0.81} & 53.06\pmv{0.68} & $-0.11$ \\
                  & ATLAS-LPF    & 52.81\pmv{0.37} & 52.60\pmv{0.73} & $-0.21$ \\
\addlinespace[2pt]
Flickr            & ATLAS        & 51.56\pmv{0.14} & 51.56\pmv{0.09} & $+0.00$ \\
                  & ATLAS-LPF-NF & 52.80\pmv{0.17} & 52.62\pmv{0.12} & $-0.18$ \\
\addlinespace[2pt]
Reddit            & ATLAS        & 95.74\pmv{0.04} & 95.77\pmv{0.05} & $+0.03$ \\
                  & ATLAS-LPF    & 96.65\pmv{0.04} & 96.20\pmv{0.01} & $-0.45$ \\
\bottomrule
\end{tabular}
\end{table}

\paragraph{Protocol.}
Under \emph{random seeding}, every Louvain call draws a fresh random
seed, so each of the $10$ runs builds an independent family $\mathcal{H}$ with
its own $|\Gamma|$. Under \emph{fixed seeding}, a single seed is reused for
every Louvain call in every run, so the community features are identical across
runs and the reported spread reflects only network initialization, dropout, and
the data split; the random-seeded spread carries both sources. Everything
downstream is unchanged under either protocol: the projection weights
$W^{(\gamma)}$ are refit from scratch in every run.

\paragraph{Results.}
Table~\ref{tab:seed} contrasts the two protocols on six benchmarks spanning
high and low structural bias. Reseeding moves accuracy by at most $1.59$ points
and by $0.33$ at the median; nine of the twelve random-seeded means lie within
one fixed-seed standard deviation of their counterpart, and the run-to-run
spread is comparable under both protocols. 

\paragraph{Discussion.}
We do not estimate $\nu^{\ast}$ directly; what Table~\ref{tab:seed} shows is
that the accuracy consequences of reseeding are small and structured as the
bounds predict. Theorem~\ref{thm:seedvi} caps the movement in the label
information a family carries, and Corollary~\ref{cor:seedrisk} carries that cap
through to expected risk---which is why accuracy, and not only information, is
what stays put. Corollary~\ref{cor:seedincl} then explains how $|\Gamma|$ can
vary at fixed accuracy: a resolution on which two seeds disagree has
information gain within $\nu^{\ast}$ of its estimation cost, so it sits near
the inclusion boundary and contributes little either way. Sensitivity is thus
confined to graphs where the community channel is retained and is not redundant
with the local channels---Corollary~\ref{cor:seedlocal} caps it independently
of $\nu^{\ast}$ once those are informative---and vanishes when the channel is
dropped. Channel selection is robust to Louvain's randomness.

\subsection{Choice of Community Detector}
\label{app:detector}

\paragraph{Detector agnosticism.}
ATLAS operates on a family of node-to-block assignments
$\mathcal{H}=\{C^{(1)},\dots,C^{(T)}\}$ at increasing granularity and is
independent of the algorithm that produces it. The one-hot encoding, the
learnable projection, concatenation with node attributes, so any detector returning a multi-scale
partition family can be substituted.

\paragraph{Why modularity-based detection.}
Modularity-based detection is a convenient mechanism for generating such
a family. Its resolution parameter is continuous, so granularity is
varied without specifying the number of communities in advance, and all
partitions are scored under a single objective, making the modularity
$Q^{(\gamma)}$ and the gap $\Delta Q(\gamma_1,\gamma_2)$ comparable
across resolutions. The adaptive search (Algorithm ~\ref{alg:adaptive-res}) uses
this comparability to choose where to evaluate next, which keeps the
number of detector calls small relative to a dense sweep; Louvain scales
to the graphs in our evaluation at a one-time cost of $O(T\|A\|_0)$
(Table ~\ref{tab:complexity}). The detector only proposes candidates: whether a
partition is retained is decided by (Theorem ~\ref{thm:2}) , since a partition
may be well formed and still add nothing beyond $X$. Alternatives are
compatible with the same pipeline under a different enumeration
procedure --- METIS requires the number of partitions to be specified a
priori~\cite{karypis1998metis}.

Table~\ref{tab:detector} substitutes Leiden and METIS for Louvain in the
community-extraction step, holding the resolution search, feature
construction, MLP, and hyperparameters fixed.%

 
\begin{table}[t]
\centering
\caption{ATLAS accuracy under three community detectors, all other
components fixed. MLP is the same backbone without the community
channel; the footer reports the mean fraction of the Louvain
community-channel gain over MLP that each detector recovers. Metrics
follow the main text (ROC--AUC on Tolokers, micro-F1 on Flickr and
Reddit, accuracy elsewhere).}
\label{tab:detector}
\scriptsize
\setlength{\tabcolsep}{3pt}
\begin{tabular}{@{}lcccc@{}}
\toprule
Dataset & MLP & Louvain & Leiden & METIS \\
\midrule
Cora            & 75.44 & \textbf{88.37$\pm$0.63} & 87.35$\pm$1.38 & 84.35$\pm$1.30 \\
Tolokers        & 72.97 & \textbf{82.19$\pm$0.73} & 81.96$\pm$0.64 & 80.51$\pm$1.15 \\
Amazon-Ratings  & 39.83 & \textbf{53.17$\pm$0.81} & 52.84$\pm$0.37 & 51.34$\pm$0.84 \\
Flickr          & 47.17 & 51.56$\pm$0.14 & 51.05$\pm$0.14 & \textbf{51.67$\pm$0.20} \\
ogbn-products   & 61.06 & \textbf{79.98$\pm$0.14} & 79.48$\pm$0.39 & 78.62$\pm$0.12 \\
Reddit          & 74.35 & \textbf{95.74$\pm$0.04} & 95.23$\pm$0.11 & 95.24$\pm$0.06 \\
\midrule
\multicolumn{2}{@{}l}{Gain retained (mean)} & 100\% & 95\% & 88\% \\
\bottomrule
\end{tabular}
\end{table}

The results are broadly consistent across detectors. All three improve
over MLP on every graph. Leiden retains $95\%$ of the Louvain
community-channel gain on average and never trails by more than $1.02$
points, while METIS retains $88\%$.
Leiden is regarded as more stable and avoids the badly-connected-community
issue of Louvain~\cite{traag2019leiden}. The small performance gap between
the two detectors suggests that ATLAS is not strongly dependent on the
specific optimization heuristic used to construct the community family. 
The detector influences the
quality of discovered partitions and the efficiency of scale exploration,
but not the applicability of the ATLAS framework. We therefore adopt
Louvain as the default implementation due to its scalability and its
compatibility with efficient multi-scale search, rather than because the
framework depends on any particular detector.

%
\subsection{Algorithms}
\label{sec:algorithms}

\begin{algorithm}[]
\caption{Adaptive Resolution Search for Louvain}
\label{alg:adaptive-res}
\begin{algorithmic}[1]
\Require graph $G$, minimum modularity $Q_{\min}$, maximum modularity gap $\Delta_{\max}$, \texttt{gap\_range}$=[a,b]$
\Ensure resolution set $\Gamma$, \texttt{community\_list}
\State $\mathcal{C}\gets\emptyset$;\ $Q\gets\emptyset$
\For{$\gamma\in\{0.5,1.0\}$} \Comment{initial resolutions}
  \State $(\mathcal{C}[\gamma],Q[\gamma])\gets\textsc{Louvain}(G,\gamma)$
\EndFor
\While{true}
  \State $L\gets\textsc{SortedKeys}(Q)$;\ $\gamma_{\max}\gets L[-1]$;\ $\tau \gets Q[\gamma_{\max}]$
  \If{$\tau\le Q_{\min}$} \State \textbf{break} \EndIf
  \State $\gamma_{\text{new}} \gets \texttt{None}$
  \For{consecutive $(\gamma_1,\gamma_2)\in L$}
     \If{$|Q[\gamma_2]-Q[\gamma_1]|>\Delta_{\max}$}
        \State $\gamma_{\text{new}} \gets (\gamma_1+\gamma_2)/2$;\ \textbf{break} \Comment{interpolate}
     \EndIf
  \EndFor
  \If{$\gamma_{\text{new}}=\texttt{None}$} \Comment{extrapolate}
     \State sample $\delta\sim\mathcal{U}[a,b]$;\ $Q^\star\gets \tau-\delta$
     \State $s\gets\textsc{EstimateSlope}(Q\text{ vs }\gamma)$
     \State $\gamma_{\text{new}} \gets \gamma_{\max}+\dfrac{Q^\star-\tau}{s}$
  \EndIf
  \State $(\mathcal{C}[\gamma_{\text{new}}],Q[\gamma_{\text{new}}])\gets\textsc{Louvain}(G,\gamma_{\text{new}})$
\EndWhile
\State $\Gamma \gets \{\, \gamma\in\textsc{SortedKeys}(Q)\!:\ Q[\gamma]\ge Q_{\min} \,\}$
\State \texttt{community\_list} $\gets [\, \mathcal{C}[\gamma]\ \text{ for } \gamma\in\Gamma\, ]$
\State \Return $\Gamma$, \texttt{community\_list}
\end{algorithmic}
\end{algorithm}

\begin{algorithm}[H]
\caption{Community-Augmented Feature Projection for Node Classification}
\label{alg:community_aug}
\begin{algorithmic}[1]
\Require Graph $G=(V,E)$, node features $\mathbf{X}\in\mathbb{R}^{n\times D}$, resolution set $\Gamma=\{\gamma_1,\ldots,\gamma_T\}$, projection dimension $d_c$
\Ensure Predicted label distribution $\hat{\mathbf{Y}}\in\mathbb{R}^{n\times C}$

\State Initialize empty list of embeddings $\mathcal{E}_{\mathrm{emb}}\gets[\,]$

\For{$\gamma\in\Gamma$}
  \State Compute community assignment $\mathbf{c}^{(\gamma)}\in\mathbb{N}^n$
  \State One-hot encode $\mathbf{c}^{(\gamma)}$: $O^{(\gamma)}\in\{0,1\}^{n\times k_\gamma}$
  \State Project via trainable weights: $\mathbf{E}^{(\gamma)}\gets \mathbf{O}^{(\gamma)}\mathbf{W}^{(\gamma)}$, where $\mathbf{W}^{(\gamma)}\in\mathbb{R}^{k_\gamma\times d_c}$
  \State Append $\mathbf{E}^{(\gamma)}$ to $\mathcal{E}_{\mathrm{emb}}$
\EndFor

\State Concatenate all embeddings: $\mathbf{E}\gets \operatorname{Concat}(\mathcal{E}_{\mathrm{emb}})\in\mathbb{R}^{n\times (T\cdot d_c)}$
\State Concatenate with node features: $\mathbf{Z}\gets \big[\mathbf{X}\,\Vert\,\mathbf{E}\big]\in\mathbb{R}^{n\times (D+T\cdot d_c)}$
\State Predict logits with MLP: $\mathbf{Y}\gets f_\theta(\mathbf{Z})\in\mathbb{R}^{n\times C}$
\State Apply softmax: $\hat{\mathbf{Y}}\gets \operatorname{softmax}(\mathbf{Y})$

\State \textbf{return}~$\widehat{\mathbf{Y}}$
\end{algorithmic}
\end{algorithm}

\subsection{Computation time on large graphs}

Table~\ref{tab:runtime} summarizes ATLAS's runtime profile on
million-node benchmarks by separating the one-time preprocessing cost
(multi-resolution community detection) from the per-epoch training and
inference costs. For ogbn-products, we additionally report ClusterGCN
and GraphSAINT under the same hardware and protocol, as requested by
Reviewers AURE and 21kS; comparable timings were not collected for the
remaining four datasets and are left as future work.

\begin{table}[t]
\centering
\footnotesize
\setlength{\tabcolsep}{3.5pt}
\renewcommand{\arraystretch}{1.1}

\caption{Preprocessing (community detection / graph partitioning),
per-epoch training, and inference times (seconds), mean $\pm$ std over
runs where available. ClusterGCN and GraphSAINT are reported only for
ogbn-products.}
\label{tab:runtime}

\begin{tabularx}{\columnwidth}{
>{\raggedright\arraybackslash}p{1.51cm}
>{\raggedright\arraybackslash}X
>{\raggedleft\arraybackslash}X
>{\raggedleft\arraybackslash}X
>{\raggedleft\arraybackslash}X
}
\toprule
Dataset & Method & Preproc. & Train / epoch & Infer. \\
\midrule

Reddit          & ATLAS      & $84.904 \pm 2.764$   & $0.143 \pm 0.002$  & $0.150 \pm 0.005$ \\
Flickr          & ATLAS      & $6.800 \pm 1.741$    & $0.241 \pm 0.005$  & $0.056 \pm 0.012$ \\
Yelp            & ATLAS      & $15.842 \pm 0.007$   & $2.670 \pm 0.007$  & $1.613 \pm 0.016$ \\
Amazon-Products & ATLAS      & $72.270 \pm 1.409$   & $6.073 \pm 0.039$  & $3.056 \pm 0.019$ \\

\midrule

Ogbn-Products & ATLAS      & $391.894 \pm 14.387$ & $0.399 \pm 0.0008$ & $0.526 \pm 0.0038$ \\
              & ClusterGCN & $168.754$            & $4.017$            & $82.837$ \\
              & GraphSAINT & $3.770^{a}$          & $0.751$            & $66.445$ \\

\bottomrule
\end{tabularx}

\vspace{2pt}
\footnotesize
\noindent $^{a}$GraphSAINT resamples subgraphs every epoch; the reported
value is per-epoch sampling cost (not directly comparable to one-time preprocessing).
\end{table}

On ogbn-products, ATLAS inference is over $150\times$ faster than both
baselines ($0.526$\,s vs.\ $82.837$\,s for ClusterGCN and $66.445$\,s
for GraphSAINT), a direct consequence of adjacency-free inference.
ATLAS's one-time preprocessing cost ($391.894$\,s) exceeds ClusterGCN's
partitioning cost ($168.754$\,s); this is an acknowledged limitation,
with GPU-accelerated multi-resolution Louvain identified as a future
optimization target. Preprocessing is $O(T\lVert A\rVert_0)$ with no
algorithmic CPU dependency, making it a tractable target for such
optimization.

\newpage

\subsection{NMI Analysis for datasets}

\begin{figure*}[!htb]
    \centering

    \begin{minipage}{0.48\textwidth}
        \centering
        \includegraphics[width=\textwidth]{plots/cora_two_panel.eps}
        \caption*{(a) Cora}
    \end{minipage}
    \hfill
    \begin{minipage}{0.48\textwidth}
        \centering
        \includegraphics[width=\textwidth]{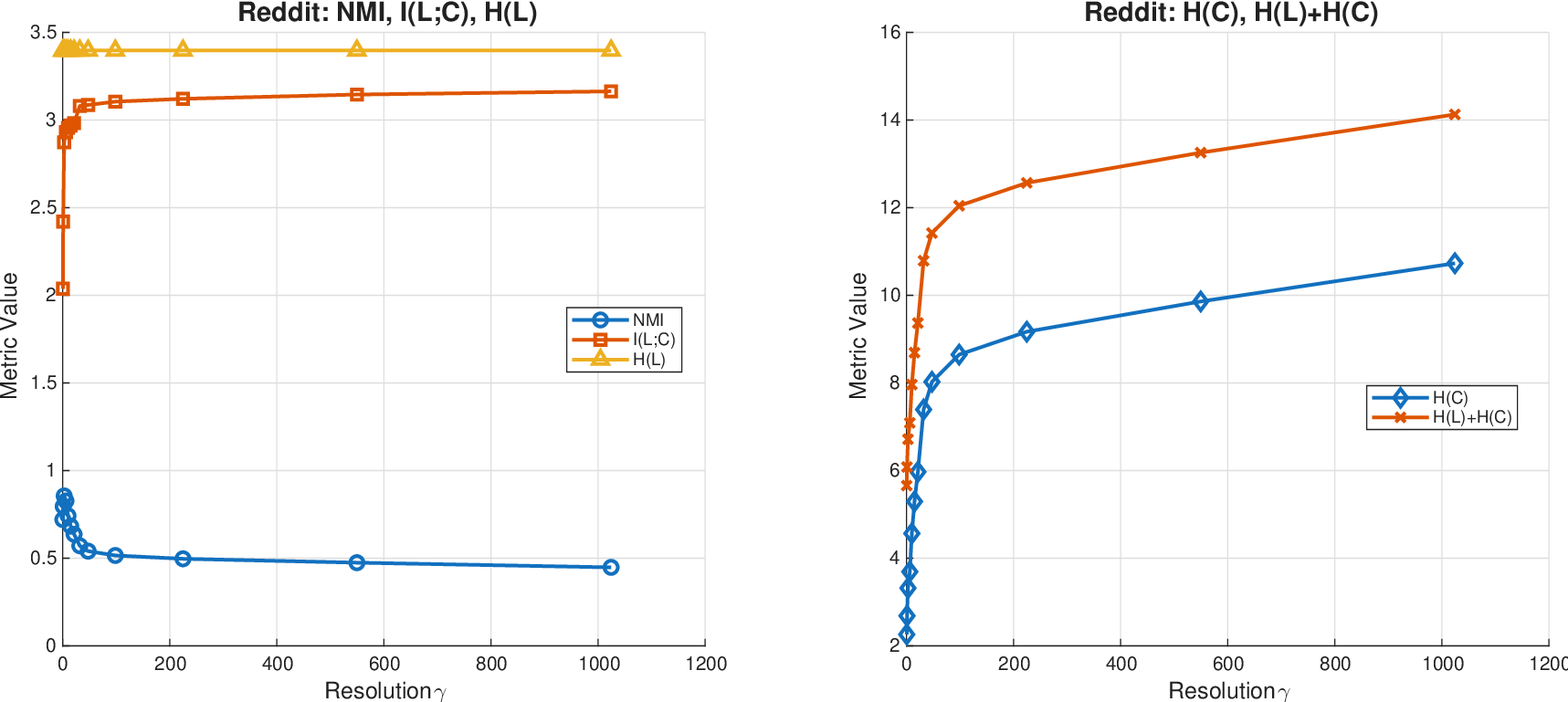}
        \caption*{(b) Reddit}
    \end{minipage}

    \vspace{0.25cm}

    \begin{minipage}{0.48\textwidth}
        \centering
        \includegraphics[width=\textwidth]{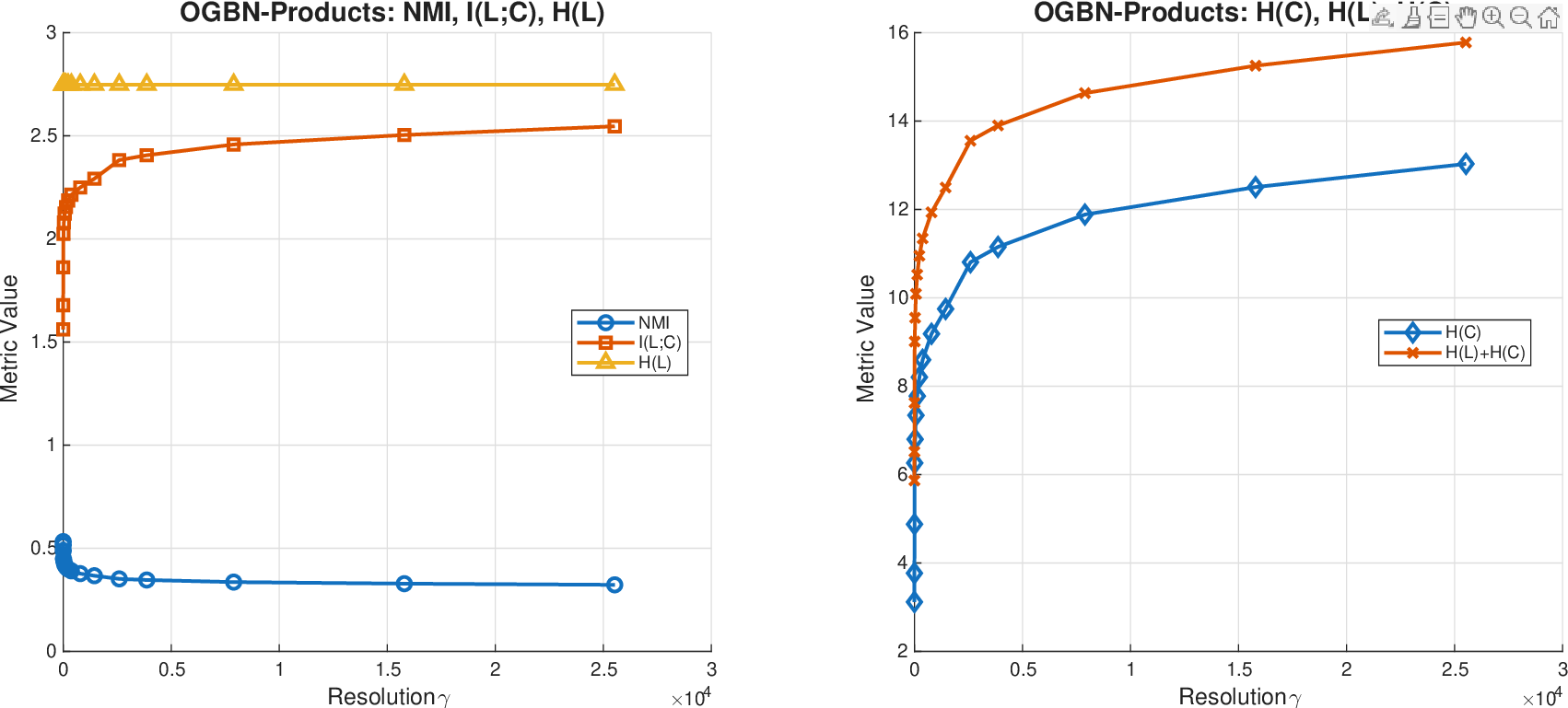}
        \caption*{(c) OGBN-Products}
    \end{minipage}
    \hfill
    \begin{minipage}{0.48\textwidth}
        \centering
        \includegraphics[width=\textwidth]{plots/tolokers_two_panel.eps}
        \caption*{(d) Tolokers}
    \end{minipage}

    \caption{
    High structural bias datasets:
    (a) Cora, (b) Reddit, (c) OGBN-Products, and (d) Tolokers.
    Each subfigure reports how $NMI$, $I(L;C)$, $H(L)$, $H(C)$, and $H(L){+}H(C)$ vary
    as the resolution parameter $\gamma$ varies and yields community partitions of different granularity.}
    \label{fig:four_datasets_multiresolution}
\end{figure*}

\begin{figure*}[t]
    \centering

    \begin{minipage}{0.48\textwidth}
        \centering
        \includegraphics[width=\textwidth]{plots/amazon_ratings_two_panel.eps}
        \caption*{(a) Amazon-Ratings}
    \end{minipage}
    \hfill
    \begin{minipage}{0.48\textwidth}
        \centering
        \includegraphics[width=\textwidth]{plots/chameleon_filtered_two_panel.eps}
        \caption*{(b) Chameleon-Filtered}
    \end{minipage}

    \vspace{0.25cm}

    \begin{minipage}{0.48\textwidth}
        \centering
        \includegraphics[width=\textwidth]{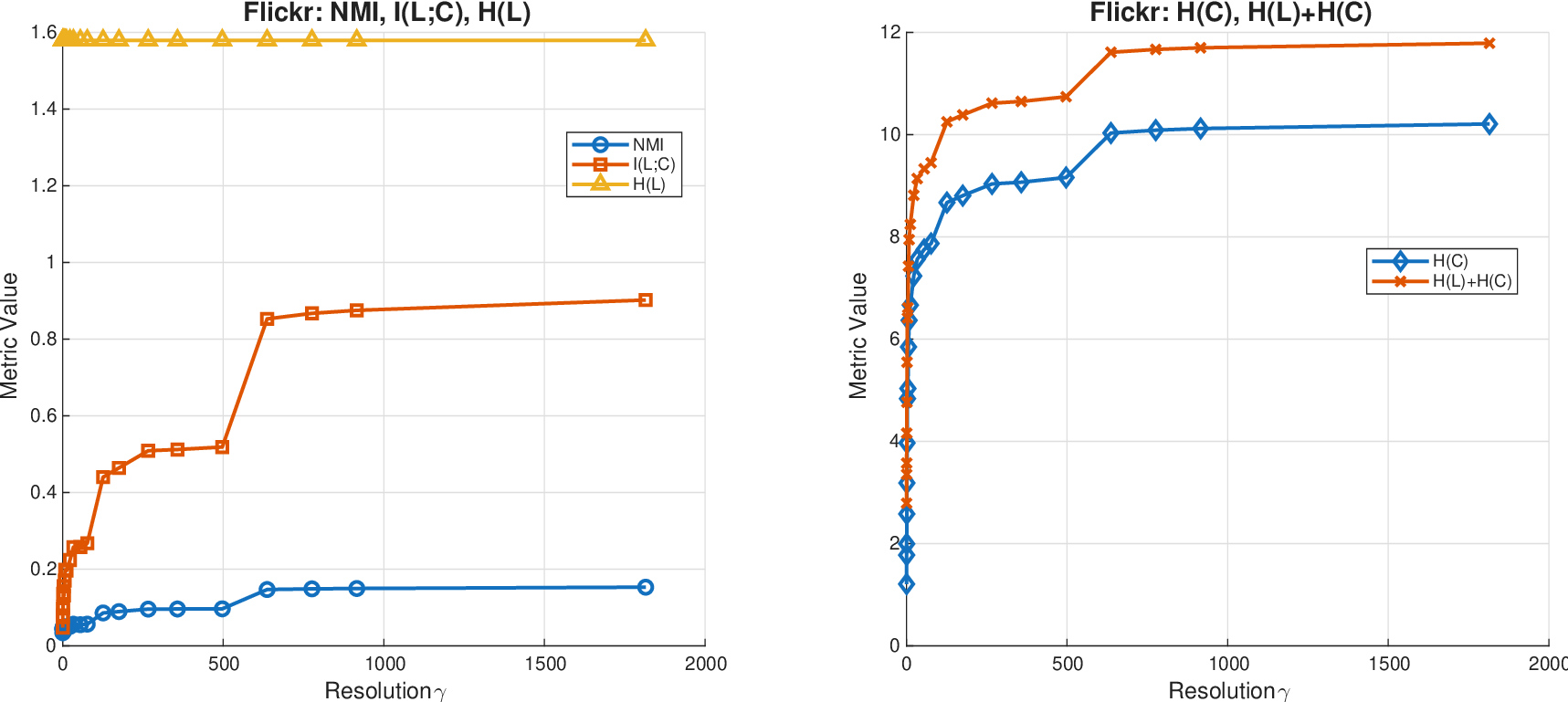}
        \caption*{(c) Flickr}
    \end{minipage}
    \hfill
    \begin{minipage}{0.48\textwidth}
        \centering
        \includegraphics[width=\textwidth]{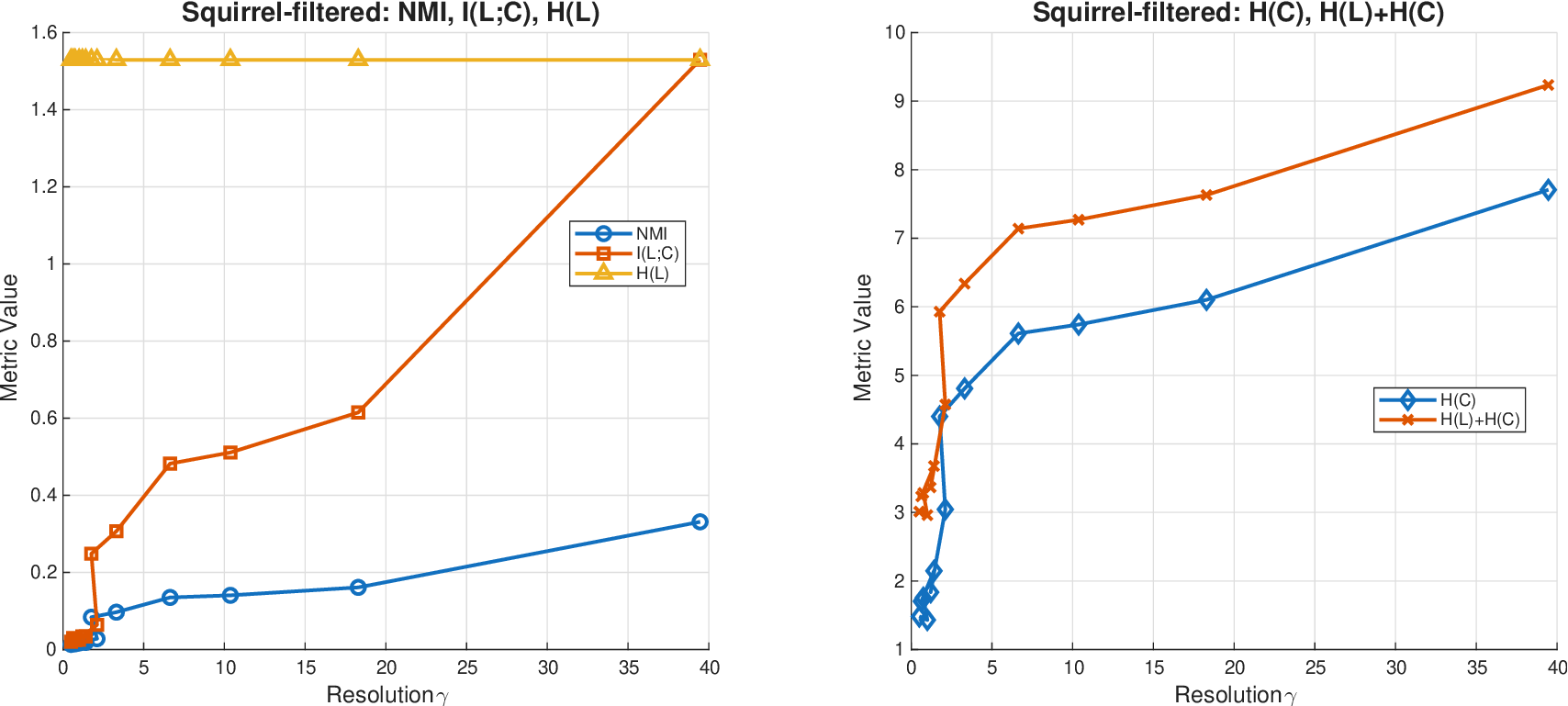}
        \caption*{(d) Squirrel-Filtered}
    \end{minipage}
    \caption{
    Low structural bias datasets:
    (a) Amazon-Ratings, (b) Chameleon-Filtered, and (c) Flickr (d) Squirrel-Filtered.
    Each subfigure illustrates how $NMI$, $I(L;C)$, $H(L)$, $H(C)$, and $H(L){+}H(C)$ evolve
    as the resolution parameter $\gamma$ varies and yields community partitions of different granularity.}
    \label{fig:low_structural_bias_datasets}
\end{figure*}

\begin{figure*}[t]
    \centering
    
    \begin{minipage}{0.48\textwidth}
        \centering
        \includegraphics[width=\textwidth]{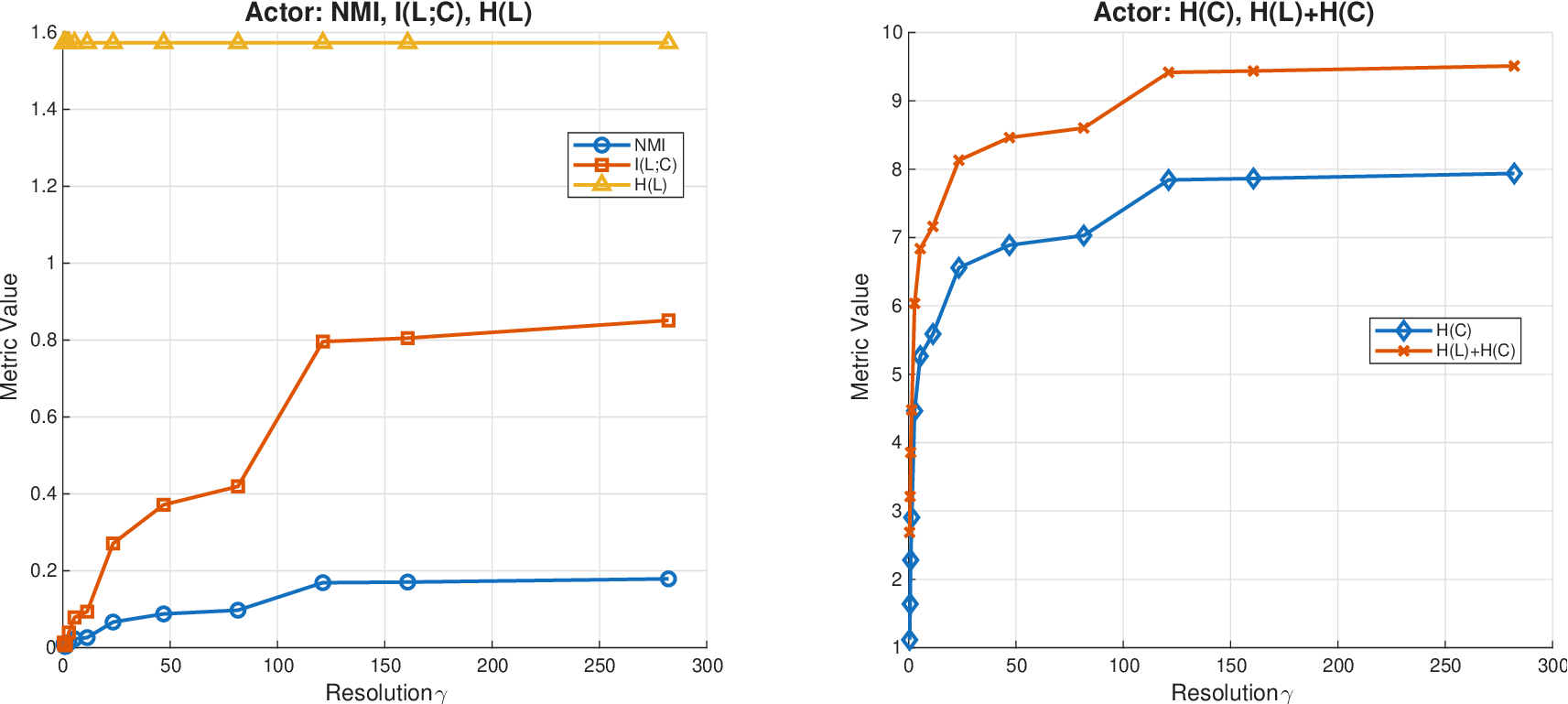}
        \caption*{(a) Actor}
    \end{minipage}
    \hfill
    \begin{minipage}{0.48\textwidth}
        \centering
        \includegraphics[width=\textwidth]{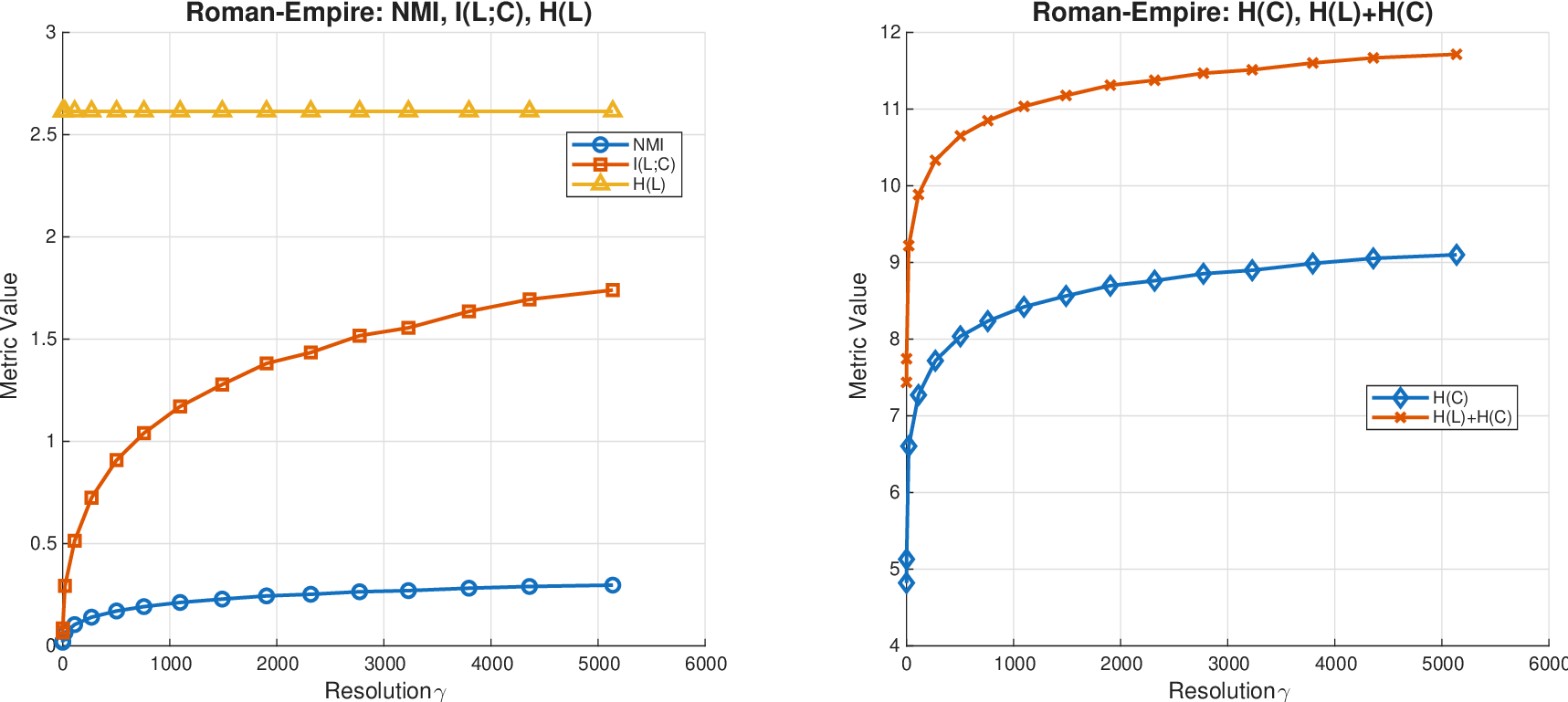}
        \caption*{(b) Roman-Empire}
    \end{minipage}

    \caption{
    Negative structural bias datasets:
    (a) Actor and (b) Roman-Empire.
    Each subfigure reports how $NMI$, $I(L;C)$, $H(L)$, $H(C)$, and $H(L){+}H(C)$ vary
    as the resolution parameter $\gamma$ varies and yields community partitions of different granularity.}
    \label{fig:negative_structural_bias_datasets}
\end{figure*}

\newpage
\subsection{Models for homophilic graphs}
\label{sec:homophily-models}

\noindent\textbf{GCN.}
Applies a linear map followed by aggregation with the symmetrically normalized adjacency (after adding self-loops), corresponding to a first-order spectral/Chebyshev approximation~\citep{kipf2017semi}.

\noindent\textbf{GAT.}
Learns attention coefficients over neighbors via masked self-attention and aggregates them with a softmax-weighted sum, enabling data-dependent receptive fields~\citep{velickovic2018graph}.

\noindent\textbf{GraphSAGE.}
Performs permutation-invariant neighbor aggregation (e.g., mean, max-pooling, LSTM) with fixed fan-out sampling per layer for scalable, inductive mini-batch training on large graphs~\citep{hamilton2017inductive}.

\subsection{Models for heterophilic graphs}
\label{sec:heterophily-models}

\noindent\textbf{H\textsubscript{2}GCN.}
Separates ego and neighbor embeddings, aggregates higher-order neighborhoods, and combines intermediate representations to improve robustness under heterophily~\citep{zhu2020beyond}.

\noindent\textbf{LINKX.}
Separately embeds node features and adjacency (structural) information with MLPs and concatenates them, capturing complementary attribute and topology signals that scale to non-homophilous graphs~\citep{lim2021linkx}.

\noindent\textbf{GPR\mbox{-}GNN.}
Learns signed polynomial (Generalized PageRank) propagation weights, adapting the filter to both homophilous and heterophilous label patterns and mitigating over-smoothing~\citep{chien2022adaptive}.

\noindent\textbf{FSGNN.}
Applies soft selection over hop-wise aggregated features with ``hop-normalization,'' effectively decoupling aggregation depth from message passing for a simple, shallow baseline that performs well under heterophily~\citep{maurya2022fsgnn}.

\noindent\textbf{GloGNN.}
Augments propagation with learnable correlations to global nodes (including signed coefficients), enabling long-range information flow and improved grouping on heterophilous graphs~\citep{li2022finding}.

\noindent\textbf{FAGCN.}
Uses a self-gating, frequency-adaptive mechanism to balance low- and high-frequency components during message passing, improving robustness across homophily regimes~\citep{bo2021beyond}.

\noindent\textbf{GBK\mbox{-}GNN.}
Employs bi-kernel feature transformations with a gating mechanism to integrate homophily- and heterophily-sensitive signals within a single architecture~\citep{du2022gbkgnn}.

\noindent\textbf{JacobiConv.}
Adopts an orthogonal Jacobi-polynomial spectral basis (often without nonlinearities) to learn flexible filters suited to varying graph signal densities, yielding strong performance on heterophilous data~\citep{wang2022jacobiconv}.


\noindent\textbf{ACM\mbox{-}GCN.}
Uses high-pass filtering with adaptive channel mixing to combine
low- and high-frequency components, yielding strong performance on heterophilic and
mixed-regime graphs~\citep{luan2022revisiting}.

\noindent\textbf{OrderedGNN.}
Orders multi-hop message passing with a gating mechanism to better control how information from different hop distances is mixed, aiming to handle heterophily while mitigating over-smoothing in deep propagation~\citep{song2023ordered}.

\noindent\textbf{M2M\mbox{-}GNN.}
Uses a \emph{multiset-to-multiset} message passing design that partitions neighbor information into multiple subsets (instead of collapsing all neighbors into a single vector), reducing harmful mixing among heterophilous neighbors and improving robustness to over-smoothing~\citep{liang2024sign}.

\noindent\textbf{CMGNN.}
Builds on a heterophilous message passing view where performance is tied to a class \emph{compatibility matrix}; CMGNN explicitly leverages and improves this compatibility structure to enhance message passing on heterophilic graphs~\citep{zheng2025cmgnn}.

\subsection{Sampling methods for scalable GNNs}
\label{sec:sampling-methods}

\noindent\textbf{GraphSAGE.}
Samples a fixed fan-out of neighbors per layer and learns permutation-invariant aggregators, limiting the receptive field and enabling inductive, mini-batch training on large graphs~\citep{hamilton2017inductive}.



\noindent\textbf{ClusterGCN.}
Partitions the graph and samples dense clusters as mini-batches, restricting propagation within blocks to boost edge coverage, cache locality, and memory efficiency at scale~\citep{chiang2019cluster}.

\noindent\textbf{GraphSAINT.}
Constructs mini-batches by sampling subgraphs (node/edge/random-walk policies) and applies unbiased normalization to correct sampling bias, yielding strong accuracy--efficiency trade-offs on large graphs~\citep{zeng2020graphsaint}.

\noindent\textbf{LABOR.}
A scalable mini-batch sampling algorithm that combines \emph{neighbor sampling} and \emph{layer sampling} via Poisson sampling. LABOR is designed as a drop-in replacement for fixed-fanout neighbor sampling: it correlates sampling across seed nodes to exploit overlap in their sampled neighborhoods, substantially reducing the number of sampled vertices/edges and mitigating neighborhood explosion while maintaining the per-node estimator variance properties of standard neighbor sampling~\citep{balin2023labor}.

\noindent\textbf{GRAPES.}
An adaptive, layer-wise neighbor sampling method for scalable mini-batch GNN training: a sampler GNN predicts node inclusion probabilities conditioned on node features/structure and the current sampled subgraph, then selects a fixed-size set of neighbors (e.g., via a Top-$k$ sampling trick) and is trained jointly with the classifier using the downstream classification loss as a learning signal (via RL/GFlowNet-style estimators), improving accuracy under small sampling budgets~\citep{younesian2025grapes}.

\subsection{Decoupling-based methods for scalable GNNs}
\label{sec:decoupling-methods}

\noindent\textbf{SGC.}
Simplifies GCNs by collapsing multiple message-passing layers into a single $K$-step precomputation of $\mathbf{A}^K\mathbf{X}$, removing nonlinearities and train-time propagation. This reduces GNN training to logistic regression on pre-smoothed features, yielding strong scalability and fast inference~\citep{wu2019simplifying}.

\noindent\textbf{SIGN.}
Precomputes multiple graph-diffused feature channels (e.g., $\mathbf{A}^K\mathbf{X}$ for several $K$), and trains an MLP on the concatenated features. This decouples feature propagation from learning entirely, enabling embarrassingly parallel preprocessing and large-batch training~\citep{rossi2020sign}.

\noindent\textbf{SAGN.}
Replaces SIGN's fixed concatenation of hop-wise features with an attention mechanism, allowing each node to adaptively weight information gathered from different hop distances rather than committing to a single receptive field. Propagation remains a one-time preprocessing step, so training proceeds in standard minibatches; a Self-Label-Enhance (SLE) stage further combines self-training with label propagation through a scalable node-label module, iteratively enlarging the labeled set across training stages~\citep{sun2021sagn}.

\noindent\textbf{GAMLP.}
Precomputes multi-hop diffused features and learns node-wise \emph{receptive-field attention} over them (recursive, JK, and GCN attention variants), so each node adaptively selects its own propagation depth; combined with reliable label utilization, it attains strong accuracy on billion-edge benchmarks without message passing at train or inference time~\citep{zhang2022gamlp}.

\medskip
\noindent
Together, these methods represent the broader ``decoupling'' paradigm---where propagation is performed once (or analytically) and training reduces to learning an MLP over fixed multi-hop representations---an approach systematically benchmarked and analyzed in large-scale settings by \citet{duan2022comprehensive}. ATLAS aligns with this propagation-free philosophy but differs fundamentally in how structural information is obtained: instead of precomputing $\mathbf{A}^k\mathbf{X}$, ATLAS extracts \emph{multi-resolution community assignments} as topology-aware features, providing a complementary and scalable route to structural encoding.

\subsection{Datasets}
\label{sec:datasets}

We evaluate on three groups of benchmarks that stress complementary regimes.

{\bf Large-scale graphs.}
We use Flickr, Reddit, Yelp, Amazon-Products, and ogbn-products. Flickr/Yelp/Amazon-Products come from GraphSAINT; Reddit from GraphSAGE; ogbn-products from OGB~\citep{zeng2020graphsaint,hamilton2017inductive,hu2020ogb}. Table~\ref{tab:exp-dataset} reports sizes, features, classes, and splits.

{\bf  Homophilous and heterophilous graphs.}
We include Cora, Questions, Chameleon-Filtered, Squirrel-Filtered, Amazon-Ratings, Tolokers, and Roman-Empire. For the filtered Wikipedia, Roman-Empire, Amazon-Ratings, Tolokers, and Questions datasets, we use the exact settings and splits of \citet{platonov2023critical}; Cora and Actor follow standard preprocessing~\citep{pei2020geomgcn,lim2021linkx}. Table~\ref{tab:exp-dataset-homophily} lists summary stats, edge homophily $h_e$, and metrics.

\begin{table}[!ht]
\centering
\caption{Dataset statistics with edge homophily $h_e$ and evaluation metric (``Acc'' for Accuracy, ``ROC-AUC'' for Area Under ROC).}
\label{tab:exp-dataset-homophily}

\resizebox{\linewidth}{!}{%
\begin{tabular}{lcccccccc}
\toprule
Dataset & Nodes & Edges & Avg. Degree & Feature & Classes & Train / Val / Test & $h_e$ & Metric \\
\midrule\midrule
Cora                & 2{,}708  & 5{,}429   & 4  & 1{,}433 & 7 (s)  & 0.60 / 0.20 / 0.20$^\dagger$ & 0.810 & Acc \\
Actor               & 7{,}600  & 30{,}019  & 8  & 932     & 5 (s)  & 0.60 / 0.20 / 0.20 & 0.216 & Acc \\
Questions           & 48{,}921 & 153{,}540 & 6  & 301     & 2 (s)  & 0.50 / 0.25 / 0.25           & 0.840 & ROC-AUC \\
Squirrel-Filtered   & 2{,}223  & 65{,}718  & 59 & 2{,}089 & 5 (s)  & 0.50 / 0.25 / 0.25           & 0.207 & Acc \\
Chameleon-Filtered  & 890      & 13{,}584  & 31 & 2{,}325 & 5 (s)  & 0.50 / 0.25 / 0.25           & 0.236 & Acc \\
Amazon-Ratings      & 24{,}492 & 93{,}050  & 8  & 300     & 5 (s)  & 0.50 / 0.25 / 0.25           & 0.380 & Acc \\
Tolokers            & 11{,}758 & 519{,}000 & 88 & 10      & 2 (s)  & 0.50 / 0.25 / 0.25           & 0.595 & ROC-AUC \\
Roman-Empire        & 22{,}662 & 32{,}927  & 3  & 300     & 18 (s) & 0.50 / 0.25 / 0.25           & 0.047 & Acc \\
\bottomrule
\end{tabular}%
}

\vspace{2pt}
{\footnotesize $^\dagger$ randomly sampled to match 60/20/20 and ensures from all Louvain communities.}
\end{table}

\begin{table}[!ht]
\caption{Dataset statistics (``m'' stands for \textbf{m}ulti-class classification, and ``s'' for \textbf{s}ingle-class.)}
\centering
\resizebox{\linewidth}{!}{
\begin{tabular}{rccccccc}
\toprule
Dataset & Nodes & Edges & Avg. Degree & Feature & Classes & Metric & Train / Val / Test\\
\midrule\midrule
Flickr          & 89{,}250      & 899{,}756      & 10   & 500 & 7 (s)    & F1-micro & 0.50 / 0.25 / 0.25\\
Reddit          & 232{,}965     & 11{,}606{,}919 & 50   & 602 & 41 (s)   & F1-micro & 0.66 / 0.10 / 0.24\\
Yelp            & 716{,}847     & 6{,}977{,}410  & 10   & 300 & 100 (m)  & F1-micro & 0.75 / 0.10 / 0.15\\
Amazon-Products  & 1{,}598{,}960 & 132{,}169{,}734& 83   & 200 & 107 (m)  & F1-micro & 0.85 / 0.05 / 0.10\\
ogbn-products   & 2{,}449{,}029 & 61{,}859{,}140 & 50.5 & 100 & 47 (s)   & Acc      & 0.08 / 0.02 / 0.90\\
\bottomrule
\end{tabular}}
\label{tab:exp-dataset}
\end{table}

{\bf Large non-homophilous graphs.}
 We use the six benchmarks of \citet{lim2021linkx},
which are one to three orders of magnitude larger than the standard
heterophilous datasets while remaining non-homophilous. We use their released features, labels, and $50/25/25$ splits, and follow their metric choice: ROC--AUC on genius, accuracy elsewhere. arXiv-year and snap-patents are
treated as directed; the remaining four are undirected.
Table~\ref{tab:exp-dataset-linkx} reports summary statistics.

\begin{table}[!ht]
\centering
\caption{Statistics of the non-homophilous benchmarks of \citet{lim2021linkx}. All splits are $0.50 / 0.25 / 0.25$, averaged over five
random splits; ``s'' denotes single-label classification.}
\label{tab:exp-dataset-linkx}
\resizebox{\linewidth}{!}{%
\begin{tabular}{lcccccc}
\toprule
Dataset & Nodes & Edges & Feature & Classes & $h_e$ & Metric \\
\midrule\midrule
Penn94                  & 41{,}554      & 1{,}362{,}229  & 5   & 2 (s) & 0.470 & Acc \\
Pokec                   & 1{,}632{,}803 & 30{,}622{,}564 & 65  & 2 (s) & 0.445 & Acc \\
arXiv-year              & 169{,}343     & 1{,}166{,}243  & 128 & 5 (s) & 0.222 & Acc \\
snap-patents            & 2{,}923{,}922 & 13{,}975{,}788 & 269 & 5 (s) & 0.073 & Acc \\
genius                  & 421{,}961     & 984{,}979      & 12  & 2 (s) & 0.618 & ROC-AUC \\
twitch-gamers           & 168{,}114     & 6{,}797{,}557  & 7   & 2 (s) & 0.545 & Acc \\
\bottomrule
\end{tabular}%
}
\vspace{2pt}

\end{table}

\newpage
\subsection{Cora: Accuracy vs. Minimum Modularity Threshold}

Table~\ref{tab:minQ_pairs_with_acc} summarizes how relaxing the minimum modularity threshold $Q_{\min}$ on Cora changes both the community-derived features and the resulting accuracy. Each tuple $(Q,\ \mathrm{Resolution},\ \mathrm{Communities})$ corresponds to a Louvain run at resolution $\gamma$: $Q$ is the modularity $Q(\gamma)$, and $\mathrm{Communities}$ is the number of communities $k_\gamma$ whose assignments $c^{(\gamma)}$ are one-hot encoded into $H^{(\gamma)}$ and projected to a dense embedding $E^{(\gamma)}$ that is concatenated into the multi-resolution community feature matrix (Algorithm~2). For a given $Q_{\min}$, the row lists the cumulative set of tuples with $Q \ge Q_{\min}$: \newpair{blue tuples} are newly activated at that threshold, while \oldpair{gray tuples} persist from higher thresholds. When $Q_{\min} \ge 0.9$, no tuples qualify and the model reduces to the base MLP with accuracy $76.61\%$. As $Q_{\min}$ is lowered from $0.8$ to $0.6$, additional high-modularity, moderate-resolution community embeddings are added, and accuracy increases up to $86.50\%$. Further decreasing $Q_{\min}$ admits lower-modularity, finer resolutions with many more communities, leading to small fluctuations and a peak accuracy of $88.40\%$ at $Q_{\min}=0.1$, where a diverse mix of coarse-to-fine community features is used. Pushing $Q_{\min}$ to $0.0$ adds one very fine tuple (768 communities), which slightly degrades performance to $86.32\%$, indicating that including too many extremely fine community features eventually injects noise.

\begin{table*}[!ht]
\centering
\caption{Cora: Cumulative $(Q,\ \mathrm{Resolution},\ \mathrm{Communities})$ pairs included at each minimum modularity threshold $Q_{\min}$ (listed in run order), with accuracy. \textit{Color coding:} pairs \newpair{colored in blue} are \emph{newly added at that $Q_{\min}$}; pairs \oldpair{in gray} were added at earlier thresholds and are carried over.}
\label{tab:minQ_pairs_with_acc}
\setlength{\tabcolsep}{6pt}
\renewcommand{\arraystretch}{1.2}
\setlength{\arrayrulewidth}{0.4pt}
\begin{tabular}{|>{\centering\arraybackslash}p{0.15\textwidth}|>{\centering\arraybackslash}p{0.66\textwidth}|>{\centering\arraybackslash}p{0.10\textwidth}|}
\hline
{\scriptsize \shortstack{\textbf{Min} \textbf{Modularity}\\$Q_{\min}$}} &
{\scriptsize \shortstack{\textbf{Pairs}\\(Modularity, Resolution,Number of Communities)}} &
{\scriptsize \shortstack{\textbf{Accuracy}}} \\
\hline
1.0 & \textemdash{} & 75.44 \\
\hline
0.9 & \textemdash{} & 75.44 \\
\hline
0.8 & \newpair{$(0.8526,\ 0.500,\ 90)$}, \ \newpair{$(0.8120,\ 1.000,\ 103)$} & 79.93 \\
\hline
0.7 & \oldpair{$(0.8526,\ 0.500,\ 90)$}, \ \oldpair{$(0.8120,\ 1.000,\ 103)$}, \ \newpair{$(0.7448,\ 2.606,\ 141)$} & 83.66 \\
\hline
0.6 & \oldpair{$(0.8526,\ 0.500,\ 90)$}, \ \oldpair{$(0.8120,\ 1.000,\ 103)$}, \ \oldpair{$(0.7448,\ 2.606,\ 141)$}, \ \newpair{$(0.6841,\ 5.483,\ 170)$}, \ \newpair{$(0.6006,\ 12.374,\ 298)$} & 86.50 \\
\hline
0.5 & \oldpair{$(0.8526,\ 0.500,\ 90)$}, \ \oldpair{$(0.8120,\ 1.000,\ 103)$}, \ \oldpair{$(0.7448,\ 2.606,\ 141)$}, \ \oldpair{$(0.6841,\ 5.483,\ 170)$}, \ \oldpair{$(0.6006,\ 12.374,\ 298)$}, \ \newpair{$(0.5566,\ 20.068,\ 325)$} & 84.55 \\
\hline
0.4 & \oldpair{$(0.8526,\ 0.500,\ 90)$}, \ \oldpair{$(0.8120,\ 1.000,\ 103)$}, \ \oldpair{$(0.7448,\ 2.606,\ 141)$}, \ \oldpair{$(0.6841,\ 5.483,\ 170)$}, \ \oldpair{$(0.6006,\ 12.374,\ 298)$}, \ \oldpair{$(0.5566,\ 20.068,\ 325)$}, \ \newpair{$(0.4909,\ 32.860,\ 373)$}, \ \newpair{$(0.4231,\ 48.392,\ 430)$} & 86.15 \\
\hline
0.3 & \oldpair{$(0.8526,\ 0.500,\ 90)$}, \ \oldpair{$(0.8120,\ 1.000,\ 103)$}, \ \oldpair{$(0.7448,\ 2.606,\ 141)$}, \ \oldpair{$(0.6841,\ 5.483,\ 170)$}, \ \oldpair{$(0.6006,\ 12.374,\ 298)$}, \ \oldpair{$(0.5566,\ 20.068,\ 325)$}, \ \oldpair{$(0.4909,\ 32.860,\ 373)$}, \ \newpair{$(0.3748,\ 63.924,\ 457)$}, \ \oldpair{$(0.4231,\ 48.392,\ 430)$} & 85.26 \\
\hline
0.2 & \oldpair{$(0.8526,\ 0.500,\ 90)$}, \ \oldpair{$(0.8120,\ 1.000,\ 103)$}, \ \oldpair{$(0.7448,\ 2.606,\ 141)$}, \ \oldpair{$(0.6841,\ 5.483,\ 170)$}, \ \oldpair{$(0.6006,\ 12.374,\ 298)$}, \ \oldpair{$(0.5566,\ 20.068,\ 325)$}, \ \oldpair{$(0.4909,\ 32.860,\ 373)$}, \ \oldpair{$(0.3748,\ 63.924,\ 457)$}, \ \oldpair{$(0.4231,\ 48.392,\ 430)$}, \ \newpair{$(0.2784,\ 95.726,\ 541)$} & 82.59 \\
\hline
0.1 & \oldpair{$(0.8526,\ 0.500,\ 90)$}, \ \oldpair{$(0.8120,\ 1.000,\ 103)$}, \ \oldpair{$(0.7448,\ 2.606,\ 141)$}, \ \oldpair{$(0.6841,\ 5.483,\ 170)$}, \ \oldpair{$(0.6006,\ 12.374,\ 298)$}, \ \oldpair{$(0.5566,\ 20.068,\ 325)$}, \ \oldpair{$(0.4909,\ 32.860,\ 373)$}, \ \oldpair{$(0.3748,\ 63.924,\ 457)$}, \ \oldpair{$(0.4231,\ 48.392,\ 430)$}, \ \oldpair{$(0.2784,\ 95.726,\ 541)$}, \ \newpair{$(0.1792,\ 136.430,\ 672)$} & 88.40 \\
\hline
0.0 & \oldpair{$(0.8526,\ 0.500,\ 90)$}, \ \oldpair{$(0.8120,\ 1.000,\ 103)$}, \ \oldpair{$(0.7448,\ 2.606,\ 141)$}, \ \oldpair{$(0.6841,\ 5.483,\ 170)$}, \ \oldpair{$(0.6006,\ 12.374,\ 298)$}, \ \oldpair{$(0.5566,\ 20.068,\ 325)$}, \ \oldpair{$(0.4909,\ 32.860,\ 373)$}, \ \oldpair{$(0.3748,\ 63.924,\ 457)$}, \ \oldpair{$(0.4231,\ 48.392,\ 430)$}, \ \oldpair{$(0.2784,\ 95.726,\ 541)$}, \ \oldpair{$(0.1792,\ 136.430,\ 672)$}, \ \newpair{$(0.0958,\ 175.819,\ 768)$} & 86.32 \\
\hline
\end{tabular}
\label{tab:explaing-cora}
\end{table*}

\end{document}